\documentclass[conference]{IEEEtran}

\newif\ifIncludeAppendix
\IncludeAppendixfalse
\IncludeAppendixtrue

\ifIncludeAppendix
\IEEEoverridecommandlockouts
\fi

\usepackage{cite}
\usepackage{amsmath,amssymb,amsfonts}
\usepackage{algorithmic}
\usepackage{graphicx}
\usepackage{textcomp}
\usepackage{xcolor}

\newcommand{\Description}[1]{}
\newcommand{\shortcite}[1]{\cite{#1}}
\newenvironment{CameraRevision}{\todo{Added for camera ready}\color{blue}}{\color{black}}
\renewenvironment{CameraRevision}{}{}

\usepackage{booktabs}

\newcommand{\argmax}[0]{\ensuremath{\operatorname{argmax}}}

\usepackage{pgfplots}
\pgfplotsset{compat=1.18}
\usetikzlibrary{pgfplots.statistics}

\pgfplotsset{
    select coords between index/.style 2 args={
        x filter/.code={
            \ifnum\coordindex<#1\fi
            \ifnum\coordindex>#2\fi
        }
    },
    rshift/.style={
        xshift=\pgfkeysvalueof{/pgfplots/rshift scale}
    },
    lshift/.style={
        xshift=-\pgfkeysvalueof{/pgfplots/lshift scale}
    },
    rshift2/.style={
        xshift=\pgfkeysvalueof{/pgfplots/rshift2 scale}
    },
    lshift2/.style={
        xshift=-\pgfkeysvalueof{/pgfplots/lshift2 scale}
    },
    rshift3/.style={
        xshift=\pgfkeysvalueof{/pgfplots/rshift3 scale}
    },
    lshift3/.style={
        xshift=-\pgfkeysvalueof{/pgfplots/lshift3 scale}
    },
    rshift4/.style={
        xshift=\pgfkeysvalueof{/pgfplots/rshift4 scale}
    },
    lshift4/.style={
        xshift=-\pgfkeysvalueof{/pgfplots/lshift4 scale}
    },
    rshift2 scale/.initial=2em,
    rshift scale/.initial=1em,
    rshift3 scale/.initial=0.3em,
    lshift3 scale/.initial=0.3em,
    rshift4 scale/.initial=0.6em,
    lshift4 scale/.initial=0.6em,
    lshift scale/.initial=1em,
    lshift2 scale/.initial=2em,
}

\usepackage[textsize=tiny, textwidth=0.625in, disable]{todonotes}

\usepackage[capitalize, noabbrev]{cleveref}

\begin{document}

\title{Dynamic Vehicle Routing Problem with\\Prompt Confirmation of Advance Requests}

\author{\IEEEauthorblockN{Amutheezan Sivagnanam\IEEEauthorrefmark{1}, Ayan Mukhopadhyay\IEEEauthorrefmark{2}, Samitha Samaranayake\IEEEauthorrefmark{3}, Abhishek Dubey\IEEEauthorrefmark{4} and Aron Laszka\IEEEauthorrefmark{1}}
\IEEEauthorblockA{\IEEEauthorrefmark{1}Department of Informatics and Intelligent Systems\\
Pennsylvania State University, University Park, Pennsylvania\\
Emails: amutheezan@psu.edu and laszka@psu.edu}
\IEEEauthorblockA{\IEEEauthorrefmark{2}Department of Computer Science\\
College of William \& Mary, Williamsburg, Virginia\\
Email: amukhopadhyay@wm.edu}
\IEEEauthorblockA{\IEEEauthorrefmark{3}School of Civil and Environmental Engineering\\
Cornell University, Ithaca, New York\\
Email: samitha@cornell.edu}
\IEEEauthorblockA{\IEEEauthorrefmark{4}College of Connected Computing\\
Vanderbilt University, Nashville, Tennessee\\
Email: abhishek.dubey@vanderbilt.edu}}

\ifIncludeAppendix
\IEEEaftertitletext{\centering Published in the Proceedings of the 17th ACM/IEEE International Conference on Cyber-Physical Systems (ICCPS 2026).}
\fi

\maketitle

\begin{abstract}
Transit agencies that operate on-demand transportation services have to respond to trip requests from passengers in real time, which involves solving \textit{dynamic vehicle routing problems with pick-up and drop-off constraints}. Based on discussions with public transit agencies, we observe a real-world problem that is not addressed by prior work: when trips are booked in advance (e.g., trip requests arrive a few hours in advance of their requested pick-up times), the agency needs to \textit{promptly confirm whether a request can be accepted or not}, and ensure that accepted requests are served as promised. State-of-the-art computational approaches either provide prompt confirmation but lack the ability to \textit{continually optimize and improve routes for accepted requests}, or they provide continual optimization but cannot guarantee serving all accepted requests. To address this gap, we introduce a novel problem formulation of \textit{dynamic vehicle routing with prompt confirmation and continual optimization}. We propose a novel computational approach for this vehicle routing problem, which integrates a quick insertion search for prompt confirmation with an anytime algorithm for continual optimization. To maximize the number requests served, we train a non-myopic objective function using reinforcement learning, which guides both the insertion and the anytime algorithms towards optimal, non-myopic solutions. We evaluate our computational approach on a real-world microtransit dataset from a public transit agency in the U.S., demonstrating that our proposed approach provides prompt confirmation while significantly increasing the number of requests served compared to existing approaches.
\end{abstract}

\begin{IEEEkeywords}
Dynamic Vehicle Routing Problem, Public Transportation, Microtransit Service, Anytime Algorithm, Machine Learning
\end{IEEEkeywords}

\newcommand{\Locations}[0]{\mathcal{L}}
\newcommand{\Request}[0]{T}
\newcommand{\Requests}[0]{\mathcal{\Request}_t}
\newcommand{\RequestDistribution}[0]{\textbf{D}}
\newcommand{\EachRequest}[0]{\Request \in \Requests}
\newcommand{\AcceptedRequests}[0]{\mathcal{\Request}^{\textit{accept}}}
\newcommand{\ActiveAcceptedRequests}[0]{\mathcal{\Request}^{\textit{accept}}_{t}}
\newcommand{\NextActiveAcceptedRequests}[0]{\mathcal{\Request}^{\textit{accept}}_{t+1}}
\newcommand{\TravelTime}[0]{D}
\newcommand{\Vehicles}[0]{\mathcal{V}}
\newcommand{\MaxCapacity}[0]{{C}}

\newcommand{\VehiclePositions}[0]{\mathcal{P}}
\newcommand{\EachVehicle}[0]{v \in \mathcal{V}}
\newcommand{\States}[0]{\mathcal{S}}
\newcommand{\TimeSymbol}[0]{\tau}
\newcommand{\LocationSymbol}[0]{l}
\newcommand{\RoutePlanSymbol}[0]{R}
\newcommand{\AbstractRoutePlans}[0]{\mathcal{\RoutePlanSymbol}}
\newcommand{\AbstractSetRoutePlans}[0]{\textbf{\RoutePlanSymbol}}
\newcommand{\RoutePlans}[0]{\AbstractRoutePlans_t}
\newcommand{\ImmediateRoutePlans}[0]{\AbstractRoutePlans^{\textit{post}}_t}
\newcommand{\NextRoutePlans}[0]{\AbstractRoutePlans^{\textit{pre}}_{t+1}}
\newcommand{\SetRoutePlans}[0]{\AbstractSetRoutePlans_t}
\newcommand{\SetImmediateRoutePlans}[0]{\AbstractSetRoutePlans^{\textit{post}}_t}
\newcommand{\RewardFunction}[0]{\textbf{r}}
\newcommand{\SimplePolicy}[0]{\pi^0}
\newcommand{\OptimalPolicy}[0]{\pi^*}
\newcommand{\IdleTimeFeature}[0]{\boldsymbol{x}}
\newcommand{\AvailabilityFeature}[0]{\boldsymbol{x}_{w}}
\newcommand{\AggregatedSpatialAvailabilityFeature}[0]{\boldsymbol{x}_{a}}
\newcommand{\AggregatedSpatialAvailabilityFeatureScalar}[0]{\chi_{a}}
\newcommand{\GridSize}[0]{\textit{g}}
\newcommand{\ContinuousInterval}[0]{N_{ci}}
\newcommand{\MinimumVehicle}[0]{N_{mv}}
\newcommand{\WindowMultiple}[0]{N_{wm}}

\section{Introduction}

The Dynamic Vehicle Routing Problem (DVRP) 
is a sequential decision-making problem that models the operation of an on-de\-mand transportation service, which serves requests for transportation between requested locations within requested time windows using a given set of vehicles  with limited passenger capacities (e.g., ride-sharing, paratransit, or microtransit services). In a DVRP, requests are received sequentially while the vehicles are in operation, serving requests that were received and accepted earlier.
When a request is received,
acceptance and routing decisions have to be made under uncertainty since future requests are stochastic, i.e., only the distribution of future requests is known.
A natural objective for this problem is to maximize the service rate, i.e., to maximize the number of requests that are accepted and served within the requested time windows. %

Existing computational approaches for this problem can be divided into two categories.
One category of algorithms decides whether to accept or reject a request when the request is received, and immediately assigns an accepted request to a vehicle manifest~\cite{alonso2017demand,wilbur2022online,joe2020deep}, removing any uncertainty for the passengers.
The other category of algorithms delays the confirmation of acceptance to continuously optimize the assignments as new requests arrive~\cite{vcivilis2023managing,karami2020periodic}, increasing the malleability of vehicle manifests, which can lead to higher service rates.
There is a gap between these two categories: an approach that provides \textit{prompt confirmation of acceptance or rejection for every request}, while also enabling the \textit{continuous improvement of vehicle manifests}. Both of these requirements are important from a practical perspective.
On the one hand, people are more likely to use a service that provides prompt confirmation, since it enables them to plan ahead and be certain that they will reach their destination on time if their requests are accepted.
On the other hand, the ability to continuously optimize assignments and vehicle manifests provides greater flexibility to service providers, which enables them to improve their service rates through optimization.
In many practical applications, especially in emerging on-demand microtransit services, these advantages are crucial.

To address this gap, we propose a novel approach that combines a quick search algorithm for request confirmation with an anytime algorithm for improving vehicle manifests \textit{between} the arrival of requests, a period of time when existing computational approaches are idle.
The quick search algorithm provides prompt confirmation, accepting or rejecting each request within a fraction of a second, which improves usability for passengers.
Between the arrival of consecutive requests, the anytime algorithm continuously works on finding better manifests, which increases services rate.
A key question faced by this approach is choosing an appropriate \textit{objective function for the anytime algorithm}.
While maximizing service rate may seem like a trivial choice for this objective, it is actually meaningless since the set of confirmed (i.e., accepted) requests is fixed between the arrival of consecutive requests.
In fact, this objective function must be non-myopic: it must maximize the chance of accepting future requests and future improvements to manifests, thereby maximizing service rate on the long term.

To address this question, we formulate the problem of request confirmation and manifest optimization as a sequential decision-making problem under uncertainty, specifically, as a Markov decision process (MDP).
The state of this MDP represents the state of the transportation service, i.e., vehicle locations, passengers on board each vehicle, set of accepted requests, current vehicle manifests, and the most recently received trip request.
An action represents the choice of accepting or rejecting the most recent request (by the search algorithm) as well as the choice of the new manifests (by the anytime algorithm).
We apply reinforcement learning to approximate the action-value function of the optimal policy in this MDP, which is the objective function that maximizes service rate in the long term in our model.
We demonstrate using real-world and synthetic problem instances that our approach provides significantly better trade-off between confirmation time and service rate than existing approaches.

The remainder of this paper is organized as follows.
\cref{sec:model} describes our formal model of the dynamic VRP with prompt request confirmation and continual manifest optimization.
\cref{sec:approach} introduces our proposed computational approach that integrates a quick search with an anytime algorithm and trains their objective function using reinforcement learning.
\cref{sec:results} presents numerical results, demonstrating that our approach achieves higher service rates with lower confirmation times compared to existing approaches.
\cref{sec:related} discusses prior work on dynamic VRP with stochastic requests.
Finally, \cref{sec:concl} provides concluding remarks.

\section{Model and Problem Formulation}
\label{sec:model}

\begin{figure*}[ht!]
    \centering
     \resizebox{\textwidth}{!}
    {\tikzset{every picture/.style={line width=0.75pt}} %

\begin{tikzpicture}
[x=0.75pt,y=0.75pt,yscale=-1,xscale=1,
    box/.style={draw, minimum width=2.5cm, minimum height=1.4cm},
    arrow/.style={->, thick}
]

\draw    (1,38) -- (658,37) ;
\draw [shift={(660,37)}, rotate = 179.91] [color={rgb, 255:red, 0; green, 0; blue, 0 }  ][line width=0.75]    (4.37,-1.32) .. controls (2.78,-0.56) and (1.32,-0.12) .. (0,0) .. controls (1.32,0.12) and (2.78,0.56) .. (4.37,1.32)   ;
\draw    (123.33,18.33) -- (123.04,34) ;
\draw [shift={(123,36)}, rotate = 271.08] [color={rgb, 255:red, 0; green, 0; blue, 0 }  ][line width=0.75]    (4.37,-1.32) .. controls (2.78,-0.56) and (1.32,-0.12) .. (0,0) .. controls (1.32,0.12) and (2.78,0.56) .. (4.37,1.32)   ;
\draw    (123,36) .. controls (123.49,65.07) and (153.59,65.65) .. (154.95,37.74) ;
\draw [shift={(155,36)}, rotate = 90.32] [color={rgb, 255:red, 0; green, 0; blue, 0 }  ][line width=0.75]    (4.37,-1.32) .. controls (2.78,-0.56) and (1.32,-0.12) .. (0,0) .. controls (1.32,0.12) and (2.78,0.56) .. (4.37,1.32)   ;
\draw    (155,36) -- (155,21) ;
\draw [shift={(155,19)}, rotate = 90] [color={rgb, 255:red, 0; green, 0; blue, 0 }  ][line width=0.75]    (4.37,-1.32) .. controls (2.78,-0.56) and (1.32,-0.12) .. (0,0) .. controls (1.32,0.12) and (2.78,0.56) .. (4.37,1.32)   ;
\draw    (437,36) .. controls (437.98,59.4) and (467.47,59.99) .. (468.94,37.75) ;
\draw [shift={(469,36)}, rotate = 90] [color={rgb, 255:red, 0; green, 0; blue, 0 }  ][line width=0.75]    (4.37,-1.32) .. controls (2.78,-0.56) and (1.32,-0.12) .. (0,0) .. controls (1.32,0.12) and (2.78,0.56) .. (4.37,1.32)   ;
\draw    (160.83,38.67) .. controls (160.67,60.67) and (392.55,63.58) .. (430.41,39.14) ;
\draw [shift={(432,38)}, rotate = 141.56] [color={rgb, 255:red, 0; green, 0; blue, 0 }  ][line width=0.75]    (4.37,-1.32) .. controls (2.78,-0.56) and (1.32,-0.12) .. (0,0) .. controls (1.32,0.12) and (2.78,0.56) .. (4.37,1.32)   ;
\draw    (472,38) .. controls (472.33,53.1) and (606.22,63.03) .. (638.91,39.43) ;
\draw [shift={(640.33,38.33)}, rotate = 140.19] [color={rgb, 255:red, 0; green, 0; blue, 0 }  ][line width=0.75]    (4.37,-1.32) .. controls (2.78,-0.56) and (1.32,-0.12) .. (0,0) .. controls (1.32,0.12) and (2.78,0.56) .. (4.37,1.32)   ;
\draw  [fill={rgb, 255:red, 184; green, 233; blue, 134 }  ,fill opacity=1 ] (104.33,3) -- (174.33,3) -- (174.33,18) -- (104.33,18) -- cycle ;

\draw  [fill={rgb, 255:red, 184; green, 233; blue, 134 }  ,fill opacity=1 ] (418.5,3) -- (488.5,3) -- (488.5,18) -- (418.5,18) -- cycle ;

\draw  [draw opacity=0] (52,77) -- (193,77) -- (193,117) -- (52,117) -- cycle ; \draw   (52,77) -- (52,117)(72,77) -- (72,117)(92,77) -- (92,117)(112,77) -- (112,117)(132,77) -- (132,117)(152,77) -- (152,117)(172,77) -- (172,117)(192,77) -- (192,117) ; \draw   (52,77) -- (193,77)(52,97) -- (193,97)(52,117) -- (193,117) ; \draw    ;
\draw  [draw opacity=0] (239,77) -- (380,77) -- (380,117) -- (239,117) -- cycle ; \draw   (239,77) -- (239,117)(259,77) -- (259,117)(279,77) -- (279,117)(299,77) -- (299,117)(319,77) -- (319,117)(339,77) -- (339,117)(359,77) -- (359,117)(379,77) -- (379,117) ; \draw   (239,77) -- (380,77)(239,97) -- (380,97)(239,117) -- (380,117) ; \draw    ;
\draw  [draw opacity=0] (427.33,77) -- (568.33,77) -- (568.33,117) -- (427.33,117) -- cycle ; \draw   (427.33,77) -- (427.33,117)(447.33,77) -- (447.33,117)(467.33,77) -- (467.33,117)(487.33,77) -- (487.33,117)(507.33,77) -- (507.33,117)(527.33,77) -- (527.33,117)(547.33,77) -- (547.33,117)(567.33,77) -- (567.33,117) ; \draw   (427.33,77) -- (568.33,77)(427.33,97) -- (568.33,97)(427.33,117) -- (568.33,117) ; \draw    ;

\draw  [fill={rgb, 255:red, 201; green, 222; blue, 244 }  ,fill opacity=1 ] (239,46.07) .. controls (239,43.64) and (240.97,41.67) .. (243.4,41.67) -- (377.27,41.67) .. controls (379.7,41.67) and (381.67,43.64) .. (381.67,46.07) -- (381.67,59.27) .. controls (381.67,61.7) and (379.7,63.67) .. (377.27,63.67) -- (243.4,63.67) .. controls (240.97,63.67) and (239,61.7) .. (239,59.27) -- cycle ;
\draw  [fill={rgb, 255:red, 155; green, 155; blue, 155 }  ,fill opacity=1 ] (197,91.92) -- (207.27,91.92) -- (207.27,88.33) -- (215.17,95.5) -- (207.27,102.67) -- (207.27,99.08) -- (197,99.08) -- cycle ;
\draw  [fill={rgb, 255:red, 155; green, 155; blue, 155 }  ,fill opacity=1 ] (384.33,91.92) -- (394.6,91.92) -- (394.6,88.33) -- (402.5,95.5) -- (394.6,102.67) -- (394.6,99.08) -- (384.33,99.08) -- cycle ;
\draw    (437.5,19) -- (437.2,34.67) ;
\draw [shift={(437.17,36.67)}, rotate = 271.08] [color={rgb, 255:red, 0; green, 0; blue, 0 }  ][line width=0.75]    (4.37,-1.32) .. controls (2.78,-0.56) and (1.32,-0.12) .. (0,0) .. controls (1.32,0.12) and (2.78,0.56) .. (4.37,1.32)   ;
\draw    (469.17,36.67) -- (469.17,21.67) ;
\draw [shift={(469.17,19.67)}, rotate = 90] [color={rgb, 255:red, 0; green, 0; blue, 0 }  ][line width=0.75]    (4.37,-1.32) .. controls (2.78,-0.56) and (1.32,-0.12) .. (0,0) .. controls (1.32,0.12) and (2.78,0.56) .. (4.37,1.32)   ;

\draw (54,80) node [anchor=north west][inner sep=0.75pt]   [align=left] {{\fontfamily{ptm}\selectfont $d_1$ \ \ \ \ \ $p_4$ $p_3$ \ \ \ \ \ $d_3$ \ $d_4$}};
\draw (74,100) node [anchor=north west][inner sep=0.75pt]   [align=left] {{\fontfamily{ptm}\selectfont $p_2$ \ \ \ \ \ \ \ \ \ $d_2$}};
\draw (241,80) node [anchor=north west][inner sep=0.75pt]   [align=left] {{\fontfamily{ptm}\selectfont $d_1$ \ \ \ \ \ $p_4$ \ \ \ \ \ \ \ \ \ \ \ \ \ $d_4$}};
\draw (260,100) node [anchor=north west][inner sep=0.75pt]   [align=left] {{\fontfamily{ptm}\selectfont $p_2$ \  $p_3$ \ \ \ \ \ $d_2$ $d_3$}};
\draw (429,80) node [anchor=north west][inner sep=0.75pt]   [align=left] {{\fontfamily{ptm}\selectfont $d_1$ \ \ \ \ \ $p_4$ \ $p_5$\ \ \ \ \ $d_5$ \ $d_4$}};
\draw (450,100) node [anchor=north west][inner sep=0.75pt]   [align=left] {{\fontfamily{ptm}\selectfont $p_2$ \ $p_3$ \ \ \ \ $d_2$ \ $d_3$}};
\draw (117.33,5) node [anchor=north west][inner sep=0.75pt]   [align=left] {{\fontfamily{ptm}\selectfont {\scriptsize \textbf{Request 4}}}};
\draw (48,22) node [anchor=north west][inner sep=0.75pt]  [font=\scriptsize] [align=left] {{\fontfamily{ptm}\selectfont request arrives}};
\draw (157,22) node [anchor=north west][inner sep=0.75pt]  [font=\scriptsize] [align=left] {{\fontfamily{ptm}\selectfont respond back with the decision}};
\draw (54,120) node [anchor=north west][inner sep=0.75pt]  [font=\scriptsize] [align=left] {{\fontfamily{ptm}\selectfont route plans after insertion of Request 4}};
\draw (241,120) node [anchor=north west][inner sep=0.75pt]  [font=\scriptsize] [align=left] {{\fontfamily{ptm}\selectfont rearranging route plans}};
\draw (429.33,120) node [anchor=north west][inner sep=0.75pt]  [font=\scriptsize] [align=left] {{\fontfamily{ptm}\selectfont route plans after insertion of Request 5}};
\draw (243.67,47.33) node [anchor=north west][inner sep=0.75pt]  [font=\scriptsize] [align=left] {{\fontfamily{ptm}\selectfont \ \ \ perform continual optimization}};
\draw (40,41.67) node [anchor=north west][inner sep=0.75pt]  [font=\scriptsize] [align=left] {{\fontfamily{ptm}\selectfont check for insertion}};
\draw (632,22) node [anchor=north west][inner sep=0.75pt]  [font=\scriptsize] [align=left] {{\fontfamily{ptm}\selectfont time}};
\draw (20,65) node [anchor=north west][inner sep=0.75pt]  [font=\scriptsize]    [align=left] {{\fontfamily{ptm}\selectfont  Vehicles}};
\draw (30,80) node [anchor=north west][inner sep=0.75pt]   [align=left] {{\fontfamily{ptm}\selectfont  $v_1$}};
\draw (30,100) node [anchor=north west][inner sep=0.75pt]   [align=left] {{\fontfamily{ptm}\selectfont $v_2$}};
\draw (208,65) node [anchor=north west][inner sep=0.75pt]  [font=\scriptsize]    [align=left] {{\fontfamily{ptm}\selectfont  Vehicles}};
\draw (218,80) node [anchor=north west][inner sep=0.75pt]   [align=left] {{\fontfamily{ptm}\selectfont $v_1$}};
\draw (218,100) node [anchor=north west][inner sep=0.75pt]   [align=left] {{\fontfamily{ptm}\selectfont $v_2$}};
\draw (397,65) node [anchor=north west][inner sep=0.75pt]  [font=\scriptsize]    [align=left] {{\fontfamily{ptm}\selectfont  Vehicles}};
\draw (407,80) node [anchor=north west][inner sep=0.75pt]   [align=left] {{\fontfamily{ptm}\selectfont $v_1$}};
\draw (407, 100) node [anchor=north west][inner sep=0.75pt]   [align=left] {{\fontfamily{ptm}\selectfont $v_2$}};
\draw (0,110) node [anchor=north west][inner sep=0.75pt]  [font=\scriptsize,rotate=-269.88] [align=left] {{\fontfamily{ptm}\selectfont {\small Route plans}}};
\draw (431.17,5) node [anchor=north west][inner sep=0.75pt]   [align=left] {{\fontfamily{ptm}\selectfont {\scriptsize \textbf{Request 5}}}};
\draw (370.83,22) node [anchor=north west][inner sep=0.75pt]  [font=\scriptsize] [align=left] {{\fontfamily{ptm}\selectfont request arrives}};
\draw (471.17,22) node [anchor=north west][inner sep=0.75pt]  [font=\scriptsize] [align=left] {{\fontfamily{ptm}\selectfont respond back with the decision}};

\end{tikzpicture}}
    \caption{During service, (1) requests arrive following a known distribution (e.g., based on historical data); (2) upon the arrival of each request, we decide whether to accept or reject the request given the current state of the service; (3) we promptly notify the passenger of our decision; (4) until the next request arrives, we continuously optimize route plans to better accommodate future requests. Symbols $p_k$ and $d_k$ represent the pickup and dropoff of request $k$, respectively.}
    \Description{During service, (1) requests arrive following a known distribution; (2) upon the arrival of each request, we decide whether to accept the request based on the current state; (3) we promptly notify the passenger of our decision; (4) until the next request arrives, we continuously optimize route plans to accommodate future requests. Symbols $p_k$ and $d_k$ represent the pickup and dropoff of request $k$, respectively.}
    \label{fig:flow_diagram}
\end{figure*}
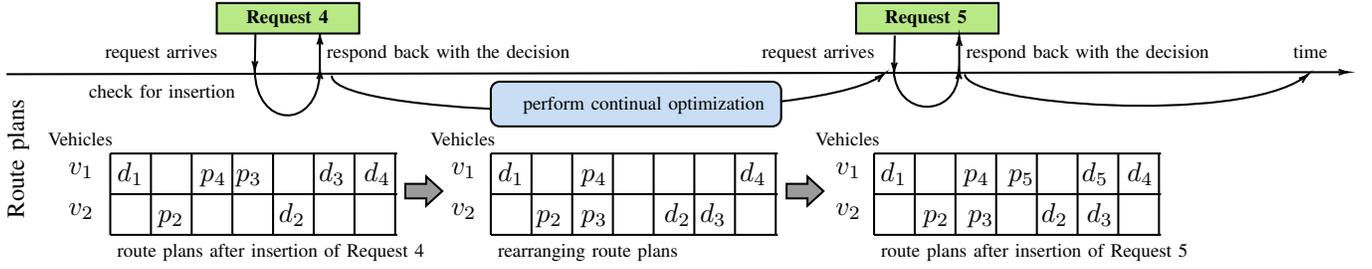

We begin by introducing our model %
of DVRP with prompt and guaranteed confirmation. %
\cref{tab:symbols} provides a list of key symbols used in the paper.

\newcommand{\hlineSwitch}[0]{} %

\begin{table}[!ht]
\centering
\renewcommand{\arraystretch}{1.1}
\setlength{\tabcolsep}{4pt}
\caption{List of Symbols}
\label{tab:symbols}
\begin{tabular}{cp{6.7cm}} %
\toprule
\textbf{Symbol} & \multicolumn{1}{c}{\textbf{Description}}  \\
\midrule
\multicolumn{2}{c}{\textbf{Variables}}  \\ 
\midrule
$\Requests$  & Set of requests that arrived as of time $t$ %
\\ \hlineSwitch
$\AcceptedRequests_t$  & Set of requests that are part of route plans at time~$t$
\\ 
\midrule
\multicolumn{2}{c}{\textbf{Constants}}  \\ \midrule
$\Locations$  & Set of locations  \\ \hlineSwitch
$\Vehicles$ & Set of vehicles \\ \hlineSwitch
$\MaxCapacity$ & Maximum passenger capacity of the vehicles \\ \hlineSwitch
$\LocationSymbol^{\textit{pickup}}_k$  & Pickup location of  request $\Request_k \in \Requests$ \\ \hlineSwitch
$\LocationSymbol^{\textit{dropoff}}_k$  & Dropoff location of  request $\Request_k \in \Requests$ \\ \hlineSwitch
$\TimeSymbol^{\textit{earliest}}_k$  & Earliest pickup time of  request $\Request_k \in \Requests$ \\ \hlineSwitch
$\TimeSymbol^{\textit{latest}}_k$  & Latest dropoff time of  request $\Request_k \in \Requests$ \\ \hlineSwitch
$\TimeSymbol^{\textit{arrival}}_k$  & Arrival time of  request $\Request_k \in \Requests$ \\ \hlineSwitch
$n_k$  & Number of passengers in request $\Request_k \in \Requests$ \\
\midrule
\multicolumn{2}{c}{\textbf{Markov Decision Process}}  \\ 
\midrule
$s_t$  & State of the environment at time $t$\\ \hlineSwitch
$\VehiclePositions_t$ & Set of vehicle locations for every vehicle $\Vehicles$ at time $t$ \\ \hlineSwitch
$\RoutePlans^{\textit{pre}}$  & Set of route plans for every vehicle $\Vehicles$ at time $t$ before making a decision \\ \hlineSwitch
$\ImmediateRoutePlans$  & Set of route plans for every vehicle $\EachVehicle$ after making a decision at time $t$ \\ \hlineSwitch
$a_t$  & Action of the decision agent at time $t$  \\ 
$\RewardFunction(s,a)$& Reward for taking action $a$ in state~$s$ \\ 
$Q(s,a)$& State-action value of taking action $a$ in state~$s$\\ 
\bottomrule
\end{tabular}
\label{table:symbol_list}
\end{table}

\subsection{Dynamic Vehicle Routing Model}
\label{sec:DVRP_model}

We let $\Locations$ denote the set of locations (e.g., pickup locations, dropoff locations, vehicle depots), which represents all points in the road network. For a pair of locations $\LocationSymbol_1,\LocationSymbol_2 \in \Locations$, we let $\TravelTime(\LocationSymbol_1,\LocationSymbol_2)$ denote the travel time from $\LocationSymbol_1$ to $\LocationSymbol_2$.
\begin{CameraRevision}
Note that our approach makes no assumptions about $D$, and it may incorporate a variety of travel-time models.
\end{CameraRevision}

Throughout the day, the transit agency receives requests one-by-one. \footnote{Multiple requests arriving at the same time can be modeled simply as sequential arrivals with zero time gap between them.}
We let $\Requests = \{T_1, T_2, \ldots\}$ denote the set of requests that have arrived up to time $t$. 
Each request $\Request_k \in \Requests$ specifies a pickup location $\LocationSymbol_{k}^{\textit{pickup}} \in \Locations$; 
a dropoff location $\LocationSymbol_{k}^{\textit{dropoff}} \in \Locations$; 
an earliest pickup time $\TimeSymbol_{k}^{\textit{earliest}} \in \mathbb{R}^{+}$; 
a latest dropoff time $\TimeSymbol_{k}^{\textit{latest}} \in \mathbb{R}^{+}$; 
and the number of passengers $n_{k} \in \mathbb{N}^{+}$. 
Each request is drawn from a known distribution $\RequestDistribution$,
i.e., $\langle \LocationSymbol_{k}^{\textit{pickup}}, \LocationSymbol_{k}^{\textit{dropoff}}, \TimeSymbol_{k}^{\textit{earliest}}, \TimeSymbol_{k}^{\textit{latest}}, n_{k} \rangle \sim \RequestDistribution$. 
As is usual in the DVRP literature, we assume that $\RequestDistribution$ can be modeled or simulated based on historical request data.
We let $\TimeSymbol_{k}^{\textit{arrival}} \in \mathbb{R}^{+}$ ($\leq \TimeSymbol_{k}^{\textit{earliest}}$) denote the arrival time of request $\Request_k \in \Requests$, i.e., the time when the transit agency receives request $\Request_k$.  %

\newcommand{\nohat}[1]{#1}

The transit agency uses a set of vehicles $\Vehicles$ to serve the requests, each vehicle having a fixed passenger capacity $\MaxCapacity \in \mathbb{N}^{+}$ (for ease of presentation, we assume uniform capacity, but our approach can be applied naturally to vehicles with heterogeneous capacities).
We let $R_{v} = \{ (\nohat{\LocationSymbol}_j, \nohat{\TimeSymbol}_j, \nohat{n}_j) \mid j =1, 2, \ldots, |R_{v}| \}$ denote the route plan (i.e., route manifest) for vehicle $\EachVehicle$, where $\nohat{\LocationSymbol}_j \in \Locations$ is either the pickup or dropoff location of a request, $\nohat{\TimeSymbol}_j \in \mathbb{R}^{+}$ is the time when vehicle $v$ will depart from location $\nohat{\LocationSymbol}_j$, and $\nohat{n}_j \in \{\pm~1,\pm~2,\ldots, \pm~\MaxCapacity\}$ is the change in the number of passengers on-board vehicle $v$ due to stopping at location $\nohat{\LocationSymbol}_j$.
\begin{CameraRevision}
Note that a valid route plan includes two stops for each request $T_k$ that is assigned to the vehicle but has not been picked up yet: a pickup stop at $\LocationSymbol_{k}^{\textit{pickup}}$ and a dropoff stop at $\LocationSymbol_{k}^{\textit{dropoff}}$ (and if the request has already been picked up, only a dropoff stop at $\LocationSymbol_{k}^{\textit{dropoff}}$).
\end{CameraRevision}
At the beginning of the day, the route plan for each vehicle $\EachVehicle$ is an empty set. 
Vehicle $\EachVehicle$ starts its shift at time $v^{\textit{start}} \in \mathbb{R}^{+}$ at the depot of the transit agency $\LocationSymbol^{\textit{depot}} \in \Locations$; it serves requests following its route plan $R_{v}$, which is updated in real time as requests are received by the transit agency; and it returns to the depot by the end of its shift $v^{\textit{end}} \in \mathbb{R}^{+}$.

\subsubsection{Operational Constraints}
\label{sec:operational_constraints}

While operating the service, the transit agency must satisfy the following constraints: 
(1) each incoming request $\Request_k \in \Requests$
must be added to either the set of accepted requests $\ActiveAcceptedRequests \subseteq \Requests$
or rejected requests $\mathcal{\Request}^{\textit{reject}}_{t} = \Requests \setminus \ActiveAcceptedRequests$;
(2) every accepted request $\Request_k \in \ActiveAcceptedRequests$ that has not been picked up yet must be assigned to exactly one vehicle, satisfying the time-window constraints (i.e., request $\Request_k$ must be picked up  after $\TimeSymbol_k^{\textit{earliest}}$ and dropped off  before $\TimeSymbol_k^{\textit{latest}}$); 
(3) every accepted request $\Request_k \in \ActiveAcceptedRequests$ that has already been picked up must keep its vehicle assignment unchanged and be dropped off  before $\TimeSymbol_k^{\textit{latest}}$; 
(4) every route plan $R_{v}$ must satisfy travel-time constraints (i.e., $\TravelTime(\nohat{\LocationSymbol}_j,\nohat{\LocationSymbol}_{j+1}) \leq \nohat{\TimeSymbol}_{j+1} - \nohat{\TimeSymbol}_j$), starting from the current location of vehicle $v$ \begin{CameraRevision}(note that this constraint implies that the stops $j$ are ordered by their departure times~${\TimeSymbol}_j$)\end{CameraRevision}; %
(5) and the number of passengers on-board vehicle $\EachVehicle$ must never exceed the passenger capacity $\MaxCapacity$.

\subsection{Computational Assumptions}
\label{sec:computer_assumptions}

When the transit agency receives a request $\Request_k \in \Requests$ at time $t = \TimeSymbol_{k}^{\textit{arrival}}$, it must decide whether to accept or reject the request \textit{within a matter of seconds}.
To accept request $\Request_k$, the agency %
must assign the request to a vehicle $\EachVehicle$ and compute a feasible route plan~$R_{v}$ for the selected vehicle that satisfies all operational constraints (\cref{sec:operational_constraints}).
Note that finding a feasible route plan is crucial to guaranteeing that the agency will be able to serve all accepted requests;
if the agency accepted a request without having a feasible route plan, it would face the risk of never finding a feasible route plan and, hence, failing to serve some of the accepted requests.
Between the arrival of two consecutive requests (i.e., between $\TimeSymbol_{k}^{\textit{arrival}}$ and $\TimeSymbol_{k+1}^{\textit{arrival}}$),
the transit agency may update the route plan of each and every vehicle, as long as the updated route plans satisfy all operational constraints (\cref{sec:operational_constraints}).
Note that the computational time available for updating route plans is stochastic since the arrival time $\TimeSymbol_{k+1}^{\textit{arrival}}$ of the next request is stochastic.

\subsection{Metrics} 
To measure the performance of the transit service up to a given time $t$, we use \textit{service rate} and \textit{rejection rate} as our key metrics, which we define as the ratio of accepted requests $|\ActiveAcceptedRequests| \, / \, |\Requests|$ and the ratio of rejected requests $|\mathcal{\Request}^{\textit{reject}}_{t}| \, / \, |\Requests|$, respectively. %

\subsection{Problem Formulation}

Our problem formulation is a \textit{two-stage optimization for a DVRP with advance requests}.
First, upon the arrival of each request, we have to promptly decide whether to accept or reject the request (and if we decide to accept the request, we have to find a feasible route plan).
Second, once we have made a decision,
we may optimize the route plans to increase the chances of accepting future requests (we may continuously optimize routes plans until the arrival of the next request).
The objective is to \textit{maximize service rate} (or, equivalently, to \textit{minimize rejection rate}) over a long time horizon ($t \rightarrow \infty$),
satisfying the operational constraints of the DVRP (\cref{sec:operational_constraints}) and the computational constraints of decision making and optimization (\cref{sec:computer_assumptions}).
\cref{fig:flow_diagram} shows an illustration of our problem setting and our proposed computational approach. %

\section{Computational Approach}
\label{sec:approach}

We propose to solve this computational problem by introducing a (1) {quick insertion algorithm} for promptly deciding whether to accept or reject a request and for finding a feasible route plan and an (2) {anytime algorithm} for continual optimization of route plans between the arrival of consecutive requests.
For both of these algorithms, we must specify an objective function, which enables us to choose between acceptance and rejection and enables us to choose between different route plans.
We propose to learn a non-myopic objective function, which maximizes service rate on the long term, using reinforcement learning.

\subsection{Markov Decision Process}

To apply reinforcement learning, we model our DVRP problem %
as a Markov decision process.

\paragraph{Decision Epoch}
A decision epoch occurs at each $t = \TimeSymbol_{k}^{\textit{arrival}}$, when request $\Request_k$ is received. 
Between the arrival of consecutive requests, the state of the environment evolves in continuous time (e.g., vehicles drive around, picking up and dropping off passengers, following their route plans). %

\paragraph{State}
The state $s_t$ of the environment at decision epoch $t = \TimeSymbol_{k}^{\textit{arrival}}$ includes
the new request $\Request_k$, the current locations $\VehiclePositions_t$ of all the vehicles,
the set of accepted requests $\ActiveAcceptedRequests$,
and the current set of feasible route plans $\RoutePlans^\textit{pre}$ for all the vehicles.

\paragraph{Action} 
The action $a_t$ taken at decision epoch $t$ includes the decision whether to \textit{accept} or \textit{reject} request $T_k$ as well as a new set of feasible route plans $\RoutePlans^\textit{post}$ for the vehicles.
An action $a_t$ can \textit{accept} the new request~$T_k$  only if the new set of feasible route plans $\RoutePlans^\textit{post}$ serves all previously accepted requests $\ActiveAcceptedRequests$ as well as the new request~$T_k$.
If action $a_t$ \textit{rejects} the new request, then the new set of feasible route plans $\RoutePlans^\textit{post}$ has to serve only the previously accepted requirements $\ActiveAcceptedRequests$
(note that in this case, the new set of route plans~$\RoutePlans^\textit{post}$ can be the same as the current set of route plans $\RoutePlans^\textit{pre}$).

\paragraph{State Transition}
At each decision epoch $t$, the environment first transitions from state $s_t$ to a post-decision state $s^\textit{post}_t$. As part of this transition, the route plans of the vehicles $\RoutePlans^\textit{pre}$ are updated to the new, post-decision route plans $\RoutePlans^\textit{post}$. If action $a_t$ \textit{accepts} the new request $T_k$, then the new request is added to the set of accepted requests: $\mathcal{T}_t^\textit{accept,post} = \ActiveAcceptedRequests \cup \{ \Request_k \}$. %
Finally, the environment transitions to its next state $s_{t'}$ upon the arrival of the next request $T_{k+1}$ at time $t' = \TimeSymbol_{k+1}^{\textit{arrival}}$.

\paragraph{Reward}
The reward is $\RewardFunction(s_t, a_t) = 1$ if action $a_t$ \textit{accepts} the new request, and $\RewardFunction(s_t, a_t) = 0$ otherwise.
Note that in the long term, maximizing rewards is equivalent to maximizing service rate (or minimizing rejection rate).

\paragraph{Action Value}
For any state-action pair $(s_t, a_t)$,
the action value $Q(s_t, a_t)$ of the optimal policy is defined as
\begin{equation}
    Q(s_t, a_t) = \RewardFunction(s_t, a_t) + \gamma \mathbb{E}_{s_{t'}} \left[\max_{a'} Q(s_{t'}, a') \right],
\end{equation}
where $\gamma \in (0, 1)$ is a temporal discount factor, and $s_{t'}$ is the stochastic next state after taking action $a_t$.
For $\gamma \approx 1$, an action $a_t$ that maximizes the action-value  $Q$ also maximizes the long-term service rate; therefore, we will employ action-value $Q$ as our objective function.
In the following subsections, we will first discuss the optimization problems of finding optimal actions for prompt confirmation (\cref{sec:prompt_confirmation}) and continual optimization (\cref{sec:continual_optimization}), and then discuss how to learn an action-value estimator (\cref{sec:RL}).

\subsection{Prompt Confirmation}
\label{sec:prompt_confirmation}

We first discuss the problem of prompt confirmation, 
which entails deciding whether to {accept} or {reject} a new request based on the current state of the environment (i.e., route plans, accepted requests, and vehicle locations), and if the decision is to accept, finding a feasible route plan that serves the new request.

\paragraph{Problem Input} The input of prompt confirmation for the decision epoch at time $t$ consists of the new request ${\Request}_k$, the current location of each vehicle $\VehiclePositions_t = \{\LocationSymbol^{v} \in \Locations \mid v \in \Vehicles \}$, the set of previously accepted requests %
$\ActiveAcceptedRequests$, and the current set of route plans~$\RoutePlans^{\textit{pre}}$ for the vehicles, which serve all the previously accepted requests~$\ActiveAcceptedRequests$.

\paragraph{Solution Space}
A solution represents the decision whether to accept or reject the new request $\Request_k$, that is, whether to include the new request $\Request_k$ in the set of accepted requests $\mathcal{T}_t^\textit{accept,post}$, and a new set of feasible route plans ${\RoutePlans^{\textit{insert}}}$ (which may be the same as $\RoutePlans^{\textit{pre}}$ if the decision is to reject the new request). %

\paragraph{Objective}
We can formally express the objective of the problem as maximizing the action-value function $Q$ of the MDP:
\begin{align}
   \max_{{\RoutePlans}^{\textit{insert}}} Q(\langle\ActiveAcceptedRequests, \Request_k, \VehiclePositions_t, \RoutePlans^{\textit{pre}}\rangle,{\RoutePlans}^{\textit{insert}} )
\end{align}
where $\langle\ActiveAcceptedRequests,  \Request_k, \VehiclePositions_t, \RoutePlans^{\textit{pre}}\rangle$ is the state at time $t$. %
Note that we omit $\mathcal{T}_t^\textit{accept,post}$ from the action representation since the inclusion (or exclusion) of request $T_k$ in the route plans ${\RoutePlans^{\textit{insert}}}$ indicates acceptance (or rejection).

\paragraph{Quick Insertion Algorithm}
To ensure that we can solve this optimization promptly, we restrict our search space for ${\RoutePlans}^{\textit{insert}}$ to \textit{searching for a simple insertion} of request $T_k$ into the current route plans ${\RoutePlans}^{\textit{pre}}$, that is, adding $T_k$ to $\RoutePlans^{\textit{pre}}$ without changing the vehicle assignments or the order of previously accepted requests.
To find a simple insertion, we can exhaustively search over each stop of each vehicle manifest $R_v \in \RoutePlans^{\textit{pre}}$, test if we can insert the pickup and dropoff stops of $T_k$ without violating the time or capacity constraints (\cref{sec:operational_constraints}), and select the feasible insertion that maximizes the $Q$ value.
\begin{CameraRevision}Since the stops of a route plan are always ordered by their departure times, the computational complexity of exhaustively searching for feasible insertions of both the pickup and dropoff stops is $O(|\ActiveAcceptedRequests|^2)$.
First, slacks in the schedule (i.e., additional time that may be spent between two consecutive stops without violating the time constraints) can be calculated by one forward and one backward pass over the route plan (linear time $O(|\ActiveAcceptedRequests|)$). Then, for each possible insertion of the pickup stop, a forward pass can check all possible insertions of the dropoff stop (quadratic time $O(|\ActiveAcceptedRequests|^2)$).
However, the search space is typically much smaller in practice.
\end{CameraRevision}
Our experimental results demonstrate that (1) we can perform an exhaustive search over all possible insertions in a fraction of a second in practice, which means that restricting the search space to insertions enables prompt confirmation, and (2) our insertion-based prompt confirmation combined with the continual optimization attains higher services rates than existing approaches.

\subsection{Continual Optimization}
\label{sec:continual_optimization}

Second, we discuss the problem of continual optimization, which entails improving route plans after each confirmation.

\paragraph{Problem Input} The input of continual optimization consists of the current location of each vehicle $\VehiclePositions_t = \{\LocationSymbol^{v} \in \Locations \mid v \in \Vehicles \}$, the set of accepted requests %
$\mathcal{T}_t^\textit{accept,post}$, and the set of feasible route plans output by the prompt confirmation $\RoutePlans^{\textit{insert}}$.

\paragraph{Solution Space}
A solution is an updated set of feasible route plans $\RoutePlans^\textit{post}$ for the vehicles, which serve all requests in $\mathcal{T}_t^\textit{accept,post}$.

\paragraph{Objective}
We can formally express the objective of the problem as maximizing the action-value function $Q$ of the MDP:
\begin{align}
 \max_{\RoutePlans^\textit{post}} Q(\langle\mathcal{T}_t^\textit{accept,post}, \VehiclePositions_t, \RoutePlans^{\textit{insert}}\rangle, \RoutePlans^\textit{post})
\end{align}
where $\langle\mathcal{T}_t^\textit{accept,post}, \VehiclePositions_t, \RoutePlans^{\textit{insert}}\rangle$ is the state of the environment at time~$t$ after the prompt confirmation.

\paragraph{Anytime Optimization Algorithm}
\label{subsec:anytime_optimization}

Solving the above problem poses two challenges.
First, similar to other formulations of vehicle routing, this problem is NP-hard~\cite{lenstra1981complexity}; even with a single vehicle, the problem is more general than the classical NP-hard Traveling Salesman Problem.
Second, we do not know in advance how much computational time $\TimeSymbol_{k+1}^{\textit{arrival}} - t$ we will have to solve the problem, since the arrival time $\TimeSymbol_{k+1}^{\textit{arrival}}$ of the next request is stochastic.
To address these challenges, we propose to apply an \textit{anytime metaheuristic algorithm}.
The advantage of employing an anytime algorithm is that we can terminate the algorithm whenever the next request~$T_{k+1}$ arrives, and we can let $\RoutePlans^\textit{post}$ be the best feasible solution found by the algorithm up that point.

\paragraph{Simulated Annealing}
We propose a \textit{simulated-annealing algorithm} as the anytime metaheuristic for continual optimization~\cite{russell2016artificial}. 
We start the simulated-annealing algorithm with $\RoutePlans^{\textit{insert}}$ as the initial feasible solution.
Starting from this initial feasible solution, the algorithm will try to improve the solution in iterations.
In each iteration of the algorithm, we perform a set of random mutation operations, including \textit{Swap}, which randomly chooses two route plans and a swappable request from each route plan (i.e., a request that is on the route but not yet picked up), and swaps these requests between the two route plans; \textit{Move}, which randomly chooses two route plans and a swappable request from one of these route plans, and moves this request to the other route plan; %
\textit{Shift}, which randomly chooses one route plan and one stop (either a pickup or dropoff) and changes the order of the stop within the route; 
and \textit{Reverse}, which selects two or more subsequent stops in a route plan and reverses their order
\begin{CameraRevision}
(note that these mutation operations are commonly used for solving VRPs~\cite{mor2022vehicle,liu2023heuristics,osman1993metastrategy,chiang1996simulated}).
\end{CameraRevision}
If the mutation results in an infeasible solution, then we revert it; if the mutation results in a feasible solution with a lower $Q$ value, then we revert it with some probability; otherwise, we continue the next iteration from the mutated solution.
When the next request $T_{k+1}$ arrives, we terminate the simulated-annealing algorithm, and we select the best feasible solution found by the algorithm up to that point as the updated set of route plans $\RoutePlans^\textit{post}$.
This way, simulated annealing acts as an anytime algorithm.

\subsection{Reinforcement Learning for Value Estimation}
\label{sec:RL}

To learn the action-value function $Q$, we employ reinforcement learning. %
During learning, we simulate the environment, and at each decision epoch $t$ (i.e., at the arrival of each request $T_k$), we find an action $a_t$ that maximizes our current estimate of function $Q$ by solving the optimization problems in \cref{sec:prompt_confirmation,sec:continual_optimization} for state~$s_t$ using the proposed algorithms.
We record the state transition as an experience, i.e., a tuple that consists of the state $s_t$ at time~$t$, the chosen action $a_t$ (i.e., decision to accept or reject $T_k$ and the updated set of route plans $\RoutePlans^\textit{post}$), the next state $s_{t'}$ (where $t' = \TimeSymbol_{k+1}^{\textit{arrival}}$), and the reward $\RewardFunction(s_t,a_t)$. 
From the recorded experiences, we can learn the action-value function $Q$ using Q-learning~\cite{watkins1992q,mnih2015human} based on the standard Bellman equation:
\begin{align}
    Q(s_t, a_t) \leftarrow \RewardFunction(s_t,a_t) + \gamma \max_{a'} Q(s_{t'}, a')
    \label{eqn:state_value_function}
\end{align}
\begin{CameraRevision}
Note that Q-learning is a value-based reinforcement learning approach, which means that there is no explicit policy $\pi^*$ mapping from states to actions; the optimal policy $\pi^*$ is implicitly defined as the solution to the maximization $\argmax_{a} Q(s_t, a)$ in prompt confirmation and continual optimization. 
\end{CameraRevision}
Next, we discuss how to design a machine learning model that can take as input a state-action pair $(s_t,a_t)$ and output an estimated~$Q$~value.

\paragraph{Feature Vectors}
Although prompt confirmation and continual optimization have different action spaces, the result of an action is always an updated set of routes, serving a set of requests. So, we can learn a single $Q$-function, whose input is a set of feasible routes (which implicitly specifies the set of accepted requests). Accordingly, we transform the complex state-action pair representation $(s_t, a_t)$  into a \textit{fixed-length feature vector} and feed it into a \textit{neural network}~$Q$.
For the neural network, we explore fully-connected feed-forward networks (MLP),
Kolmogorov-Arnold networks (KAN) \cite{liu2025kan}, 
and convolutional neural networks (CNN); 
we present experimental results for these various architectures as part of our numerical evaluation.
Here, we describe three different feature-vector representations that we can feed into the various neural network architectures. 
\ifIncludeAppendix
We present a more detailed discussion of hyperparameter search and feature selection in Appendix~\ref{app:arch_search}.
\else
We present a more detailed discussion of hyperparameter search and feature selection in Appendix~A~\cite{sivagnanam2026dynamic}.
\fi

\textbf{Total idle time:} First, we introduce a simple feature $\IdleTimeFeature$ that measures the total idle time (i.e., time when a vehicle is idle between serving two requests) for all route plans for the next $h$ hours:
\begin{align}
    \IdleTimeFeature = \sum_{R_v \in \RoutePlans} \sum_{j=0}^{|R_v|}  \textbf{I}({R_{v}}^{+}, j) 
    \label{eqn:idle_time_computation}
\end{align}
where {${R_{v}}^{+} = \{ (\nohat{\LocationSymbol}_j, \nohat{\TimeSymbol}_j, \nohat{n}_j) \mid j = 0, 1, 2, \ldots, |R_{v} + 1| \}$ denotes the complete route plan, including the start (i.e., $j = 0$) and end stops (i.e., $j = |R_{v} + 1|$) of the route, where the start stop is either the initial location (i.e., stationed at a depot) or the current location of the vehicle, and the end stop is stationed back at the depot}.  
$\textbf{I}({R_{v}}^{+}, j)$ denotes the idle time between two consecutive stops $j$ and $j+1$ in the route plan. We can formally express $\textbf{I}({R_{v}}^{+}, j)$ as follows:
\begin{align*}
    \textbf{I}&({R_{v}}^{+}, j) \!=\! \max \{ \min \{\TimeSymbol_{j + 1}, t + h \} \!-\! \TimeSymbol_{j}  \!-\! \TravelTime({\LocationSymbol_{j}, \LocationSymbol_{j + 1}}), 0 \}
\end{align*}

\textbf{Temporal availability:}
Next, we consider a feature vector~$\AvailabilityFeature$ that captures the fine-grained availability of vehicles for next $h$ hours.
Specifically, we divide the next $h$ hours into time intervals of length $w$, resulting in $\frac{h}{w}$ time intervals.
For time interval $i \in \{0, \ldots, \frac{h}{w} - 1\}$, we count the number of vehicles that are idle during time interval~$i$, and we let the feature value $({\AvailabilityFeature})_i$ be this number.

\textbf{Spatio-temporal availability:}
Finally, we introduce a feature vector $\AggregatedSpatialAvailabilityFeature$, which represents summarized vehicle availability over the next $h$ hours with fine granularity. To construct this feature, we first partition the spatial area of the service into a two-dimensional grid of size ($\GridSize \times \GridSize$). Each grid cell $[c_p, c_q]$ (where $c_p, c_q \in \{1, \ldots, \GridSize\}$) is further divided into time intervals of length $w$, yielding a total of $\frac{h}{w}$ intervals. For each interval $i \in \{0, \ldots, \frac{h}{w} - 1\}$, we count the number of idle vehicles and denote this count as $({\AggregatedSpatialAvailabilityFeatureScalar})_i^{pq}$.
Next, for each grid cell $[c_p, c_q]$, we compute the minimum number of idle vehicles in the continuous subarray of size $\WindowMultiple \times \ContinuousInterval$, where $\WindowMultiple$ represents the multiple of the window to be considered, and $\ContinuousInterval$ represents the number of such continuous intervals to be evaluated. Accordingly, we compute the feature vector for all possible combinations of $[c_p, c_q] \in \{1, \ldots, \GridSize
\} \times \{1, \ldots, \GridSize\}$, $n_{ci} \in \{1, \ldots, \ContinuousInterval\}$, and $n_{mv} \in \{1, \ldots, \MinimumVehicle\}$:
\begin{align*} \AggregatedSpatialAvailabilityFeature = \sum_{i=\kappa}^{\frac{h}{w}} \textbf{1} \mid \{\min \{({\AggregatedSpatialAvailabilityFeatureScalar})_{i -\kappa}^{pq}, \ldots, ({\AggregatedSpatialAvailabilityFeatureScalar})_{i -1}^{pq}\} \geq n_{mv} \} 
\end{align*}
where $\kappa = \WindowMultiple \times n_{ci}$.
Accordingly, at the end we obtain a feature vector of size $\GridSize \times \GridSize \times \ContinuousInterval \times \MinimumVehicle$.

\begin{CameraRevision}
The mapping from the complete state $s_t$ to a fixed-length feature vector ($\boldsymbol{x}$, $\boldsymbol{x}_w$, or $\boldsymbol{x}_a$) is necessary because the complete state representation is highly unstructured and variable length. The mapping reduces this to a structured, fixed-length vector, which can be fed into a relatively simple neural network. Alternatively, we could try to employ a complex neural architecture to ingest the complete representation. However, a complex architecture would require substantially more training experiences. Our experimental results in \cref{sec:results} demonstrate that our vector-representations are both computationally tractable and attain excellent performance.
\end{CameraRevision}

\subsection{Supervised Pre-Training}
\label{subsec:supervised_pretrain}

While reinforcement based training can lead to an optimal solution, the training process is computationally expensive. 
To tackle this challenge, we apply simple supervised learning for pre-training, where we first gather experiences by simulating the environment with %
a simple policy $\SimplePolicy$.
This simple policy $\SimplePolicy$ always accepts the new request $T_k$ if there is a feasible insertion, and it chooses action~$a_t$ by maximizing the total idle time for next $h$ hours:
\begin{align}
    \SimplePolicy(s_t) = \argmax_{{a_t}} \sum_{R_v \in \RoutePlans} \sum_{k=0}^{|R_v|}  \textbf{I}({R_{v}}^{+}, k) 
    \label{eqn:idle_time_maximization}
\end{align}
where $\RoutePlans$ is the set of route plan that we obtain from applying the action ${a_t}$ to the state $s_t$.

To apply supervised learning, we need to provide ground truth; in this case, we must provide ground truth for the action values~$Q^{\SimplePolicy}(s, a)$ of the simple policy $\pi^0$, which we will use to pre-train our neural network $Q$. 
We estimate the ground truth~$Q^{\SimplePolicy}(s, a)$ based on the discounted sum of the future rewards for next $k$ steps:
\begin{align}
    Q^{\SimplePolicy}(s_t, a_t) \approx \sum_{i=0}^{k-1} \gamma^{i} \RewardFunction(s_{t+i}, a_{t+i})
    \label{eqn:state_value_approximation}
\end{align}
This $k$-step Monte Carlo estimate of the action value $Q^{\SimplePolicy}(s, a)$ strikes a balance between minimizing bias in the estimate (due to considering a limited number of steps $k$) and minimizing the variance in the estimate (due to the stochasticity of future state transitions and rewards).

Once the supervised pre-training of our neural network~$Q$ has converged, we can apply reinforcement learning with the Bellman equation to fine-tune the policy, as described in \cref{sec:RL}.
\begin{CameraRevision}
Once reinforcement learning has converged, we can deploy our approach with the learned $Q$; during execution, we only need to solve $\argmax_{a} Q(s_t, a) $ for each decision.
\end{CameraRevision}

\section{Numerical Evaluation}
\label{sec:results}

\subsection{Datasets and Experimental Setup}

We evaluate our proposed computational approach on two datasets: a real-world microtransit dataset  from a mid-size U.S. city 
and the widely-used NYC taxi dataset.
This novel microtransit dataset and all software artifacts, including the implementation of our simulation environment and our computational framework, are available at \texttt{https://github.com/aronlaszka/DVRP-AR}

\paragraph{Microtransit Data}

We obtained real-world microtransit data from a public transit agency that serves a mid-size U.S.\ city. 
The dataset contains trip requests from August 2022 to October 2023. 
We model the arrival $\TimeSymbol_{k}^{\textit{arrival}}$ of requests $T_k$ as a Poisson process, matching the average arrival rate of the dataset (i.e., 28 requests per hour).
We model the time difference between request arrival $\TimeSymbol_{k}^{\textit{arrival}}$ and requested pickup time $\TimeSymbol_{k}^{\textit{earliest}}$ as an exponential distribution, matching the average time difference between request arrival and requested pickup time of the dataset (i.e., 2 hours). 
We consider a vehicle capacity of $c = 8$ and let the number of vehicles be $|\Vehicles| = 4$, matching the resources of the real-world service. 
We calculate travel times between all pairs of locations using real road-network data from OpenStreetMap.
The road network contains 4,555 nodes (i.e., intersection) and 11,410 edges (i.e., road segments).

\paragraph{NYC Taxi Data}
We also evaluate our approach on the NYC taxi dataset, which has been used very widely in VRP research~\cite{nycdata,kim2023rolling,alonso2017demand}. 
The NYC taxi dataset consists of trip requests from January 2016. 
We model the arrival of requests as a Poisson process with an arrival rate of 60 requests per hour. 
Since this dataset contains requests with only a short lead time (i.e., short time between request arrival and requested pickup time, since the service did not support advance booking), we adapted the dataset so that some requests are booked in advance.
Specifically, we model the time difference between request arrival and requested pickup time as an exponential distribution, matching the average time difference between request arrival and requested pickup time in the microtransit dataset  (i.e., 2 hours).
Similarly, we again consider a vehicle capacity of $c = 8$ and let the number of vehicles be $|\Vehicles| = 4$.
We again calculate travel times using real road-network data from OpenStreetMap.
The road network contains 14,512 nodes (i.e., intersection) and 33,763 edges (i.e., road segments).

\paragraph{Evaluation Episodes}
For evaluation, we consider each episode to be 1 day.
For the microtransit data, we consider 5 different episodes with 616 requests in total,
and for the NYC taxi data, we consider 5 different episodes with 1,140 requests in total.
To make the comparison fair, we evaluate our proposed approach and all the baseline approaches on the exact same 5 + 5 sets of requests.

\subsection{Baseline Approaches}

We compare our proposed solution approach to the following baseline approaches.

\begin{itemize}
\item \textbf{Google OR-Tools:} 
OR-Tools Routing is a publicly available, state-of-the-art VRP solver, developed by Google~\shortcite{ortools}.
We implemented a dynamic VRP solver based on the OR-Tools Routing API by using the \textit{Local Cheapest Cost Insertion} algorithm to make the quick insertion decision at the arrival of a request and using the \textit{Guided Local Search} algorithm to optimize the route plans between the arrival of the requests.

\item \textbf{Rolling Horizon (RH):} 
Kim \emph{et al.}~\shortcite{kim2023rolling} recently published a state-of-the-art VRP solver that supports pickups and dropoffs with time windows, based on a rolling-horizon temporal decomposition and the widely used RTV-ILP framework for dynamic VRP~\cite{alonso2017demand}.
We apply the Rolling Horizon (RH) solver by running a complete RH optimization upon the arrival of each request to determine whether we can accept the request and to optimize route plans.

\item \textbf{MC VRP:}
Wilbur \emph{et al.}~\shortcite{wilbur2022online} introduced a dynamic VRP solver that is non-myopic (i.e., considers future requests when making acceptance and route-optimization decisions) based on Monte Carlo tree search (MC VRP).
We run a full MC VRP search on the arrival of each request.
\end{itemize}

Although prior work has explored learning-based solution approaches for dynamic VRP, most existing approaches consider simplified or abstract settings, and as a result, none of them can be applied to our problem setting without substantial changes. For example,  Joe and Lau~\shortcite{joe2020deep} consider a dynamic VRP with only dropoffs (no pickups); Zhang \emph{et al.}~\shortcite{zhang2021solving} consider only a single vehicle. Due to these limitations, we cannot make a direct, fair comparison between our proposed approach and these existing learning-based approaches; we discuss the limitations of existing approaches in more detail in \cref{sec:related}.

\subsection{Implementation and Hyperparameters}

\paragraph{Software and Hardware}
We implemented our framework using Python 3.11 and Keras 3.6 with PyTorch 2.5 backend~\cite{paszke2019pytorch}. 
We ran all algorithms on a server with 2 AMD EPYC 7763 64-core CPUs, 1TB of RAM, and 2 NVIDIA RTX A5000 GPUs with 24GB of VRAM.

\paragraph{Architectures and Hyperparameters}
We represent our action-value function $Q(s, a)$ as a neural network. 
We performed an architecture search over different types of neural networks, including multi-layer perceptron (MLP), Kolmogorov-Arnold network (KAN), and convolutional neural network (CNN).
\ifIncludeAppendix
We describe these architectures and their hyperparameters in detail in  Appendix~\ref{app:arch_search}.
\else
We describe these architectures and their hyperparameters in detail in  Appendix~A~\cite{sivagnanam2026dynamic}.
\fi
For MLP, we optimized the number of layers, the number of neurons in each layer, and the drop-out rates; 
for KAN, we optimized the number of layers and the width of each layer; 
and for CNN, we optimized the number of channels, the size of the kernel, the size of the feed-forward network, and the drop-out rates.
In all cases, we used a learning rate of 0.001 for gradient descent, and a temporal discount factor of $\gamma = 0.9$.

\paragraph{Pre-Training}
For both the microtransit dataset and the NYC taxi dataset, we gathered 1 million experiences for supervised pre-training by simulating the simple policy $\pi^0$ (i.e., always accepting requests and maximizing idle time).
We performed the actual pre-training with 800,000 experiences, using the remaining 200,000 experiences as a hold-out test set. 
The trained $Q$-function achieves an $R^2$ score of 0.995 for microtransit data and 0.984 for NYC taxi data. With a batch size of 1024, each training epoch takes around 15 seconds for MLP, 33 seconds for CNN, and around 45 seconds and for KAN. On average, the supervised pre-training converged after 11 epochs (less than 3 minutes) for MLP,
after 130 epochs (around 97 minutes) for KAN, and after 22 epochs (around 12 minutes) for CNN.  Among the three neural-network architectures, MLP and KAN perform slightly better than CNN with respect to the rejection rate, on both the microtransit and the NYC taxi data. 
\ifIncludeAppendix
We provide results based on MLP models in the main text, and we provide results based on the other two architectures in Appendices~\ref{app:rejection_rate} and~\ref{app:varying_running_time}.
\else
We provide results based on MLP models in the main text, and we provide results based on the other two architectures in Appendices~B~and~C~\cite{sivagnanam2026dynamic}.
\fi

\subsection{Numerical Results}
\label{subsec:numerical_res}

\subsubsection{Running Time of Confirmation}
A key consideration is how quickly we can confirm the acceptance or rejection of a trip request.
The average {running time of our confirmation algorithm} with the trained action-value function~$Q$ %
is around \emph{0.2 seconds for microtransit data} and around \emph{1 second for NYC data}.  
This is sufficiently low to accept or reject incoming requests in real time. %
\textbf{Google OR-Tools} takes around 0.1 seconds for microtransit data and around 0.5 seconds for NYC data, which is also sufficiently low (however, it underperforms our proposed approach in terms of rejection rate, as discussed below). 
For \textbf{RH}, we let the rolling-horizon factor be 2, and we let the solver perform RTV-graph generation for up to 1 second per batch, where each batch groups requests within a pre-defined interval (e.g., 60 seconds). 
For microtransit data, solving each decision epoch using the RH approach takes 50 seconds on average, with a minimum of 7 seconds and a maximum of 122 seconds. 
\ifIncludeAppendix
In Appendix~\ref{app:rh_performance}, 
\else
In Appendix~H~\cite{sivagnanam2026dynamic}, 
\fi
we discuss the results of additional experiments where we let the RTV-graph generation take up to 5, 10, and 15 seconds with various rolling horizon factors, such as 1 and 2.
For \textbf{MC VRP}, we let the running time at each decision epoch be 30 seconds.
\ifIncludeAppendix
We provide additional experimental results with different running times, ranging from 1 to 30 seconds per decision epoch, in Appendix~\ref{app:mcvrp_performance}.
\else
We provide additional experimental results with different running times, ranging from 1 to 30 seconds per decision epoch, in Appendix~G~\cite{sivagnanam2026dynamic}.
\fi

\subsubsection{Rejection Rates}
The second key consideration is what fraction of requests we have to reject.
\cref{fig:accept_or_reject_full} shows the distribution of rejection rates over 5 different episodes from the microtransit data. Our proposed computational approach $\pi^*$ significantly outperforms the baselines, and it is able to reduce rejection rates to around 1\%.
\cref{fig:accept_or_reject_full_nyc} shows the distribution of rejection rates over 5 different episodes from NYC data. Again, our computational approach $\pi^*$ significantly outperforms all baselines. %
These results demonstrate that our proposed approach, which combines a quick insertion for confirmation with an anytime metaheuristic for continual optimization, provides by far the best trade-off:
it significantly outperforms Google OR-Tools in terms of rejection rate, and it significantly outperforms the other two baselines in terms of both rejection rate and confirmation time.

\definecolor{ColorCustomGreen}{rgb}{0, 0.8, 0}
\definecolor{ColorOlive}{rgb}{0.72, 0.71, 0.42}
\colorlet{ColorOurRL}{magenta}
\colorlet{ColorLegendOurRL}{ColorOurRL!50}
\colorlet{ColorOurRLMLP}{ColorCustomGreen}
\colorlet{ColorLegendOurRLMLP}{ColorOurRLMLP!50}
\colorlet{ColorOurRLConv}{ColorOlive}
\colorlet{ColorLegendOurRLConv}{ColorOurRLConv!50}
\colorlet{ColorOurHeu}{blue}
\colorlet{ColorLegendOurHeu}{ColorOurHeu!50}
\colorlet{ColorORTools}{red}
\colorlet{ColorLegendORTools}{ColorORTools!50}
\colorlet{ColorMCVRP}{orange}
\colorlet{ColorLegendMCVRP}{ColorMCVRP!50}
\colorlet{ColorRH1}{brown}
\colorlet{ColorLegendRH1}{ColorRH1!50}
\definecolor{ColorMaroon}{rgb}{0.5, 0, 0}
\colorlet{ColorRH2}{ColorMaroon}
\colorlet{ColorLegendRH2}{ColorRH2!50}
\colorlet{ColorOffline}{black}
\colorlet{ColorLegendOffline}{ColorOffline!50}

\pgfplotstableread[col sep=comma,]{data/carta/base/base_results.csv}\baseresults

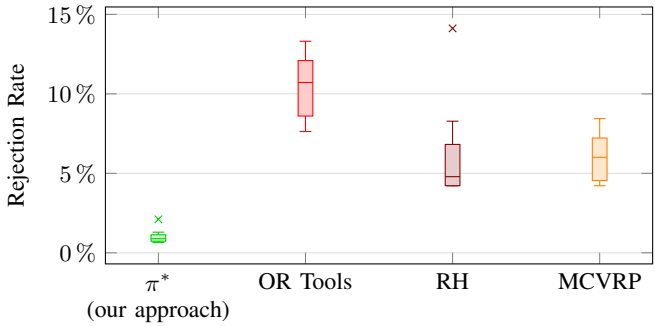
\begin{figure}[t]
\begin{tikzpicture}
\begin{axis}[
      boxplot/draw direction=y,
      width=\columnwidth,
      xtick={1,2,3,4},
      xticklabel style={align=center},
      xticklabels={{$\OptimalPolicy$\\(our approach)}, {OR Tools}, {RH}, {MCVRP}},
      height = 5cm,
      ymajorgrids,
      major grid style={draw=gray!25},
      bugsResolvedStyle/.style={},
      ylabel={Rejection Rate},
            yticklabel=\pgfmathprintnumber{\tick}\,$\%$,
      xlabel style={align=center},
      yticklabel=\pgfmathprintnumber{\tick}\,$\%$,
      font=\small
    ]

\addplot+[boxplot={box extend=0.1, draw position=1}, ColorOurRLMLP, solid, fill=ColorOurRLMLP!20, mark=x] table [col sep=comma, y=rejection_rate_with_rejection_mlp] {\baseresults};
\addplot+[boxplot={box extend=0.1, draw position=2}, ColorORTools, solid, fill=ColorORTools!20, mark=x] table [col sep=comma, y=rejection_rate_rr_it] {\baseresults};
\addplot+[boxplot={box extend=0.1, draw position=3}, ColorRH2, solid, fill=ColorRH2!20, mark=x] table [col sep=comma, y=rejection_rate_rh_2_1] {\baseresults};
\addplot+[boxplot={box extend=0.1, draw position=4}, ColorMCVRP, solid, fill=ColorMCVRP!20, mark=x] table [col sep=comma, y=rejection_rate_mc_vrp] {\baseresults};
\end{axis}
\end{tikzpicture}
\caption{Distribution of request rejection rates across 5 different episodes from the real-world microtransit data.}
\Description{Distribution of request rejection rates across 5 different episodes from the real-world microtransit data.}
\label{fig:accept_or_reject_full}
\end{figure}

\pgfplotstableread[col sep=comma,]{data/nyc/base/base_results.csv}\baseresults

\begin{figure}[t]
\begin{tikzpicture}
\begin{axis}[
      boxplot/draw direction=y,
      width=\columnwidth,
      xtick={1,2,3,4},
      xticklabel style={align=center},
      xticklabels={{$\OptimalPolicy$\\(our approach)}, {OR Tools}, {RH}, {MCVRP}},
      height = 5cm,
      ymajorgrids,
      major grid style={draw=gray!25},
      bugsResolvedStyle/.style={},
      ylabel={Rejection Rate},
            yticklabel=\pgfmathprintnumber{\tick}\,$\%$,
      xlabel style={align=center},
      font=\small
    ]

\addplot+[boxplot={box extend=0.1, draw position=1}, ColorOurRLMLP, solid, fill=ColorOurRLMLP!20, mark=x] table [col sep=comma, y=rejection_rate_with_rejection_mlp] {\baseresults};
\addplot+[boxplot={box extend=0.1, draw position=2}, ColorORTools, solid, fill=ColorORTools!20, mark=x] table [col sep=comma, y=rejection_rate_rr_it] {\baseresults};
\addplot+[boxplot={box extend=0.1, draw position=3}, ColorRH2, solid, fill=ColorRH2!20, mark=x] table [col sep=comma, y=rejection_rate_rh_2_1] {\baseresults};
\addplot+[boxplot={box extend=0.1, draw position=4}, ColorMCVRP, solid, fill=ColorMCVRP!20, mark=x] table [col sep=comma, y=rejection_rate_mc_vrp] {\baseresults};
\end{axis}
\end{tikzpicture}
\caption{Distribution of request rejection rates across 5 different episodes from the NYC taxi data.}
\Description{Distribution of request rejection rates across 5 different episodes from the NYC taxi data.}
\label{fig:accept_or_reject_full_nyc}
\end{figure}
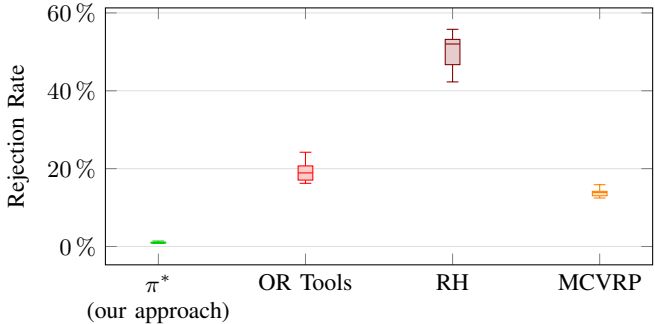

\pgfplotstableread[col sep=comma,]{data/carta/search_time/search_time_mlp.csv}\mlpsearchtime

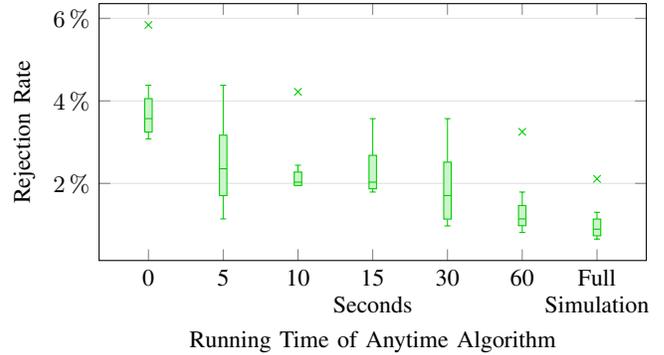
\begin{figure}[t]
\begin{tikzpicture}
\begin{axis}[
      boxplot/draw direction=y,
      width=\columnwidth,
      xtick={1,2,3,4,5,6,7},
      xticklabel style={align=center},
      xticklabels={{0}, {5}, {10}, {15\\Seconds}, {30}, {60}, {Full\\ Simulation}},
      height = 5cm,
      ymajorgrids,
      major grid style={draw=gray!25},
      bugsResolvedStyle/.style={},
      ylabel={Rejection Rate},
            yticklabel=\pgfmathprintnumber{\tick}\,$\%$,
      xlabel style={align=center},
      xlabel={Running Time of Anytime Algorithm},
      font=\small,
    ]

\addplot+[boxplot={box extend=0.1, draw position=1}, ColorOurRLMLP, solid, fill=ColorOurRLMLP!20, mark=x] table [col sep=comma, y=reject_insert_ar] {\mlpsearchtime};

\addplot+[boxplot={box extend=0.1, draw position=2}, ColorOurRLMLP, solid, fill=ColorOurRLMLP!20, mark=x] table [col sep=comma, y=reject_time_5_ar] {\mlpsearchtime};

\addplot+[boxplot={box extend=0.1, draw position=3}, ColorOurRLMLP, solid, fill=ColorOurRLMLP!20, mark=x] table [col sep=comma, y=reject_time_10_ar] {\mlpsearchtime};

\addplot+[boxplot={box extend=0.1, draw position=4}, ColorOurRLMLP, solid, fill=ColorOurRLMLP!20, mark=x] table [col sep=comma, y=reject_time_15_ar] {\mlpsearchtime};

\addplot+[boxplot={box extend=0.1, draw position=5}, ColorOurRLMLP, solid, fill=ColorOurRLMLP!20, mark=x] table [col sep=comma, y=reject_time_30_ar] {\mlpsearchtime};

\addplot+[boxplot={box extend=0.1, draw position=6}, ColorOurRLMLP, solid, fill=ColorOurRLMLP!20, mark=x] table [col sep=comma, y=reject_time_60_ar] {\mlpsearchtime};

\addplot+[boxplot={box extend=0.1, draw position=7}, ColorOurRLMLP, solid, fill=ColorOurRLMLP!20, mark=x] table [col sep=comma, y=reject_search_ar] {\mlpsearchtime};
\end{axis}
\end{tikzpicture}
\caption{Distribution of rejection rates across 5 different episodes from the real-world microtransit data, with various running-time limits for the anytime algorithm  (in seconds).} %
\Description{Distribution of rejection rates across 5 different episodes from the real-world microtransit data, with various running-time limits for our anytime algorithm  (in seconds).}
\label{fig:mlp_search_time_accept_or_reject_decision}
\end{figure}

\subsection{Ablation Studies}

Finally, we present ablation studies that investigate the benefits of continual optimization and employing the learned objective $Q$.

\subsubsection{Varying the Running Time of the Anytime Algorithm} 
First, we investigate the benefits of continual optimization, i.e., the benefit of running the anytime algorithm between the arrival of consecutive requests to improve route plans.
\cref{fig:mlp_search_time_accept_or_reject_decision} shows how the rejection rate depends on the running time of the anytime algorithm (on the microtransit data).
In this experiment, we limit the running time of the anytime algorithm to a given time allotment instead of running it until the arrival of the next request.
The results demonstrate the importance of taking advantage of the time between consecutive request arrivals since the rejection rate decreases significantly as the running time increases.

\subsubsection{Learned $Q$-Function}
Second, we investigate the benefits of employing the learned action-value function $Q$ as a non-myopic objective.
In these experiments, we replace the learned $Q$-function with the objective of the simple heuristic $\SimplePolicy$ in various parts of our framework, including the acceptance/rejection decision, the quick insertion, and the continual optimization.
\ifIncludeAppendix
We provide the results of these various ablation studies in Appendix~\ref{app:ablation}.
\else
We provide the results of these various ablation studies in Appendix~E~\cite{sivagnanam2026dynamic}.
\fi

\section{Related Work}
\label{sec:related}

Here, we provide an overview of computational approaches for dynamic VRP, focusing on recent learning-based approaches, which are most relevant to our work.
For a more detailed overview of neural combinatorial optimization for vehicle routing problems, we refer the reader to the recent survey by Wu \emph{et al.}~\shortcite{wu2024neural}.

\subsection{Supervised Learning}
\label{subsec:related_supervised}

Jungel \emph{et al.}~\shortcite{jungel2023learning} propose learning-based online optimization for large-scale online dispatching and rebalancing. %
To find online, non-myopic dispatching and rebalancing actions, the approach formulates finding a decision as a combinatorial optimization problem that is parameterized by a machine learning (ML) layer and trained using structured learning (SL). %
In contrast to our approach, Jungel et al.\ assume that rides cannot be shared.
Time is discretized into periods, and at the end of each period, the system decides which requests to accept out of requests arrived within each period. Due to this, all the requests have to wait until the end of the period for confirmation of acceptance.

Baty \emph{et al.}~\shortcite{baty2024combinatorial} consider the DVRP with Time Windows (DVRPTW), where  requests must be served by strictly satisfying time windows.
In contrast to our work, Baty et al.\ assume that there is an unlimited fleet of homogeneous vehicles and the goal is minimizing total travel costs.
They propose a ML pipeline that incorporates a combinatorial optimization layer.
This pipeline discretizes the planning horizon into a set of epochs: in each epoch, the fleet operator solves a dispatching and vehicle routing problem for the epoch-dependent request set (i.e., requests that have been received but not served yet), deciding which requests to serve in that epoch.
In both \cite{jungel2023learning,baty2024combinatorial}, once a request is assigned to a vehicle, the assignment cannot be changed. In contrast, our approach continually optimizes routes, providing a significant advantage (see \cref{fig:mlp_search_time_accept_or_reject_decision}).

\subsection{Reinforcement Learning and Neural Approximate Dynamic Programming}
\label{subsec:related_rl}

Joe and Lau~\shortcite{joe2020deep} consider the DVRP with both known and stochastic requests, formulating it as a route-based MDP.
In their model, the state of the MDP encompasses vehicle locations, manifests, and received requests; an action is updating the manifest of a vehicle; and rewards capture the objective of minimizing travel time and time-windows violations, while all the requests must be served.
In contrast, the objective of our approach is to maximize service rate, while strictly satisfying time windows.
Joe and Lau employ temporal-difference (TD) learning to learn an approximate value function of the optimal policy in this MDP %
and they propose a routing heuristic based on simulated annealing to find actions that maximize the value function.
In contrast to our work, Joe and Lau do not consider running simulated annealing between the arrival of consecutive requests; they constrain the action space so that requests are always accepted and cannot be swapped between vehicles; and they consider delivery (i.e., drop-off only) instead of pick-up and drop-off problem.

Shah \emph{et al.}~\shortcite{shah2020neural} propose a neural approximate dynamic programming approach, called NeurADP, for the on-demand ride-pooling problem.
They formulate the Ride-pool Matching Problem (RMP) with a given set of vehicles, stochastic requests, and the objective of maximizing the number of requests served.
In this formulation, requests must be served in the near future, subject to delay constraints (maximum allowed pick-up delay, maximum allowed detour delay); in contrast, our approach considers arbitrary pick-up and drop-off time windows. NeurADP divides time into decision epochs, and at the end of each epoch, it assigns a subset of the requests received in the current epoch to the vehicles; once a request is assigned to a vehicle, its assignment cannot be changed. In contrast, our approach provides prompt confirmation for requests (instead of waiting until the end of an epoch) and continual optimization of route plans (including changing assignments).

Ma \emph{et al.}~\shortcite{ma2021hierarchical} consider DVRP with both pickups and dropoffs. %
They solve the problem using hierarchal RL; when a request is received, it is first added to a buffer, and a high-level RL agent (based on DQN) decides when to release the requests.
Once released the request are assigned to vehicles and manifests are iteratively optimized by a low-level RL agent (based on REINFORCE).
In contrast to our approach, this framework runs iterative optimization intermittently (instead of continually) and for limited time, and it does not consider the choice of accepting or rejecting requests (minimizing time-window violations instead). Zhang \emph{et al.}~\shortcite{zhang2021solving} consider the Dynamic Traveling Salesman Problem (DTSP) and the Dynamic Pickup and Delivery Problem (DPDP) with stochastic, time-dependent travel times and stochastic customers (i.e., customers may be dynamically added to or deleted from the pool of customer who have not been picked up yet).
Note that this approach considers a single vehicle. Pan and Liu~\shortcite{pan2023deep} consider the Dynamic and Uncertain Vehicle Routing Problem (DU-VRP), whose objective is to serve the uncertain demands of customers in a dynamic environment.

Azagirre \emph{et al.}~\shortcite{azagirre2024better} present the RL-based platform that Lyft uses for the online rideshare matching problem, which matches trip requests with available drivers in batch.
Since matching decisions affect the distribution of available drivers (and, hence, the number of requests that the platform will be able to fulfill in the future), the matching must be non-myopic.
To this end, the platform models the online rideshare matching problem as a MDP, and it employs an online RL framework that learns and updates the value function completely online and on-policy while generating the matching decisions in real~time.

Finally, Huang \emph{et al.}~\shortcite{huang2020dynamic} propose a heuristic based solution approach for a DVRP with pre-booked and on-demand requests. However, acceptance decisions are delayed until operational hours begin, and the approach considers only single-capacity vehicles without ride-sharing.

\section{Conclusions}
\label{sec:concl}

Motivated by the emergence of on-demand microtransit services,
we introduced a novel problem formulation, the \textit{dynamic vehicle routing problem with advance booking}, and a novel computational approach
that provides prompt confirmation of trip requests while enabling the continual improvement of route plans. 
Our experimental results demonstrate that our approach can make decisions on whether to accept or reject a request \textit{in a fraction of a second} and that our anytime optimization of route plans leads to \textit{substantially lower rejection rates} than existing approaches.
The significance of these results lies in the ability to deploy on-demand microtransit services that (1) allow passengers to request trips in advance, promptly confirming the acceptance (or rejection) of their requests and guaranteeing that accepted requests will be served, and (2) achieve high service efficiency in terms of the fraction of trip requests that are accepted and served on time.

\section*{Acknowledgment}
This material is based upon work supported by the National Science Foundation (NSF) under Award No. CNS-1952011 and by the U.S. Department of Energy (DOE) under Award No. DE-EE0011188.
Any opinions, findings and conclusions or recommendations expressed in this material are those of the author(s) and do not necessarily reflect the views of the NSF and the U.S. DOE.
We thank the anonymous reviewers for their feedback on our work and for their suggestions to improve our manuscript.

\bibliographystyle{IEEEtran}
\bibliography{main}

\ifIncludeAppendix
\clearpage
\setcounter{page}{1}

\appendix

\section{Supplementary Numerical Results}

\subsection{Neural-Network Architecture Search}
\label{app:arch_search}

\renewcommand{\hlineSwitch}[0]{} %

\begin{table*}[t]
\centering
\caption{Neural-Network Hyperparameters}
\label{tab:best_architectures}
\begin{tabular}{ccc}%
\toprule
\multicolumn{3}{c}{\textbf{Neural Network Architecture: MLP}}\\ 
\midrule
\textbf{Name of the Hyperparameter} & \textbf{Microtransit Data} & \textbf{ NYC Taxi Data}\\ 
\hlineSwitch
Number of Hidden Layers & 3 & 2\\ \hlineSwitch
Number of Neurons & [64, 128, 128] & [128, 128]\\ \hlineSwitch
Dropout Rate & 0.1 & 0.2\\ 
\toprule
\multicolumn{3}{c}{\textbf{Neural Network Architecture: KAN}}\\ 
\midrule
\textbf{Name of the Hyperparameter} & \textbf{Microtransit Data} & \textbf{NYC Taxi Data}\\ 
\hlineSwitch
Number of KAN Layers & 2 & 2\\ \hlineSwitch
Layer Widths & [9, 9] & [2, 1]\\ 
\toprule
\multicolumn{3}{c}{\textbf{Neural Network Architecture: CNN}}\\ 
\midrule
\textbf{Name of the Hyperparameter} & \textbf{Microtransit Data} & \textbf{NYC Taxi Data}\\ 
\hlineSwitch
Output Channels of 1$^{st}$ Convolution Layer & 8 & 8\\ \hlineSwitch
Output Channels of 2$^{nd}$ Convolution Layer & 4 & 8\\ \hlineSwitch
Kernel Size & 3 & 3 \\ \hlineSwitch
Number of Neurons & 64 & 128\\ \hlineSwitch
Dropout Rate & 0.1 & 0.1 \\ 
\bottomrule
\end{tabular}
\end{table*}

\begin{table*}[t]
\centering
\caption{Feature Vector Combination}
\label{tab:best_feature_combination}
\begin{tabular}{ccc}%
\toprule
\multicolumn{3}{c}{\textbf{Feature Selection: MLP}}\\ 
\midrule
\textbf{Name of the Parameter} & \textbf{Microtransit Data} & \textbf{ NYC Taxi Data}\\ 
\hlineSwitch
Feature Combination & $\IdleTimeFeature,  \AggregatedSpatialAvailabilityFeature$ & $\IdleTimeFeature,  \AggregatedSpatialAvailabilityFeature$ \\ 
Grid dimension ($\GridSize$) & 1 & 3 \\
Multiple of window ($\WindowMultiple$) & 2 &  2 \\
Number of subsequent intervals ($\ContinuousInterval$)  & 1 & 1  \\
Number of idle vehicle at an interval ($\MinimumVehicle$) & 1 &  1\\
\toprule
\multicolumn{3}{c}{\textbf{Feature Selection: KAN}}\\ 
\midrule
\textbf{Name of the Parameter} & \textbf{Microtransit Data} & \textbf{NYC Taxi Data}\\ 
\hlineSwitch
Feature Combination & $\IdleTimeFeature,  \AggregatedSpatialAvailabilityFeature$ & $\IdleTimeFeature,  \AggregatedSpatialAvailabilityFeature$\\ 
Grid dimension ($\GridSize$) & 3 & 3 \\
Multiple of window ($\WindowMultiple$) & 2 &  2\\
Number of subsequent intervals ($\ContinuousInterval$) & 1 & 1\\
Number of idle vehicle at an interval ($\MinimumVehicle$) & 2 & 1 \\
\bottomrule
\end{tabular}
\end{table*}

\paragraph{MLP}
For MLP-based architecture, we optimize the number of hidden layers, the number of neurons for each hidden layer, and the dropout rate. 
We consider hyperparameters in the following search space:
\begin{itemize}
    \item Number of Hidden Layers: 1 / 2 / 3 layers
    \item Number of Neurons per Hidden Layers: 32 / 64 / 128 neurons per layer
    \item Dropout Rate (after each hidden layer): 0.0 / 0.1 / 0.2
\end{itemize}

During the hyperparameter search, we consider features based on the following two sets of features and their combinations $\IdleTimeFeature,\AggregatedSpatialAvailabilityFeature$. For $\AggregatedSpatialAvailabilityFeature$ and combination of $\IdleTimeFeature,\AggregatedSpatialAvailabilityFeature$, we consider the following search space:

\begin{itemize}
    \item Grid dimension: 1 / 2 / 3
    \item Multiple of window: 1 / 2 / 4
    \item Number of subsequent intervals: 1 / 2 / 4
    \item Number of idle vehicle at an interval: 1 / 2 / 4
\end{itemize}

On both microtransit and NYC taxi data, the MLP-based model performs better when using both features combined (see \cref{tab:best_feature_combination}).

\paragraph{KAN}
For KAN, we optimize the number of KAN layers and the width of each KAN layer, considering the following search space:
\begin{itemize}
    \item Number of KAN Layers: 1 / 2 layers
    \item Width of KAN Layers: 1 / 2 / 3 / 4 / 5 / 6 / 7 / 8 / 9
\end{itemize}

During the hyperparameter search, we consider features based on the following two sets of features and their combinations $\IdleTimeFeature,\AggregatedSpatialAvailabilityFeature$. For $\AggregatedSpatialAvailabilityFeature$ and combination of $\IdleTimeFeature,\AggregatedSpatialAvailabilityFeature$, we consider the following search space:

\begin{itemize}
    \item Grid dimension: 1 / 2 / 3
    \item Multiple of window: 1 / 2 / 4
    \item Number of subsequent intervals: 1 / 2 / 4
    \item Number of idle vehicle at an interval: 1 / 2 / 4
\end{itemize}

On both microtransit and NYC taxi data, the MLP-based model performs better when using both features combined (see \cref{tab:best_feature_combination}).

\paragraph{CNN}
For CNN, we consider two 1-dimensional convolution layers followed by a single layer of feed-forward network.
We optimize the number of output channels for each layer, the kernel size, the number of neurons in the feed-forward layer, and the dropout rate, considering the following search space:
\begin{itemize}
    \item Number of Output Channels: 4, 8, 16
    \item Kernel Size: 2, 3, 4
    \item Number of Neurons in the Feed-Forward Layer: 64 / 128 neurons
    \item Dropout Rate (after each hidden layer): 0.0 / 0.1
\end{itemize}

For CNN-based architecture, we consider a feature vector~$\AvailabilityFeature$ that captures the fine-grained availability of vehicles for next $h$ hours. In our experiments, we let $h = 4$ hours and $w = 30$ seconds, which results in a feature vector $\boldsymbol{x}$ of length 480.

\cref{tab:best_architectures} shows the best hyperparameters for each of the neural-network architectures (i.e., MLP, KAN, and CNN) for both microtransit and NYC taxi data. \cref{tab:best_feature_combination} shows the best feature combination for each neural-network architecture (i.e., MLP, KAN) for both microtransit and NYC taxi~data.

\subsection{Rejection Rates with Various Neural-Network Architectures}
\label{app:rejection_rate}

\pgfplotstableread[col sep=comma,]{data/carta/base/base_results.csv}\baseresults

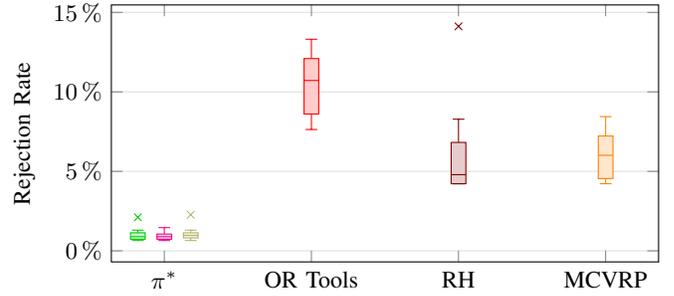
\begin{figure}
\begin{tikzpicture}
\begin{axis}[
      boxplot/draw direction=y,
      width=\columnwidth,
      xtick={1,2,3,4},
      xticklabel style={align=center},
      xticklabels={{$\pi^*$}, {OR Tools}, {RH}, {MCVRP}},
      height = 5.0cm,
      ymajorgrids,
      major grid style={draw=gray!25},
      bugsResolvedStyle/.style={},
      ylabel={Rejection Rate},
            yticklabel=\pgfmathprintnumber{\tick}\,$\%$,
      xlabel style={align=center},
      font=\small,
    ]

\addplot+[boxplot={box extend=0.1, draw position=1}, ColorOurRLMLP, solid,  lshift, fill=ColorOurRLMLP!20, mark=x] table [col sep=comma,y=rejection_rate_with_rejection_mlp] {\baseresults};
\addplot+[boxplot={box extend=0.1, draw position=1}, ColorOurRL, solid, fill=ColorOurRL!20, mark=x] table [col sep=comma, y=rejection_rate_with_rejection_kan] {\baseresults};
\addplot+[boxplot={box extend=0.1, draw position=1}, ColorOurRLConv, solid, rshift, fill=ColorOurRLConv!20, mark=x] table [col sep=comma,  y=rejection_rate_with_rejection_cnn] {\baseresults};
\addplot+[boxplot={box extend=0.1, draw position=2}, ColorORTools, solid, fill=ColorORTools!20, mark=x] table [col sep=comma, y=rejection_rate_rr_it] {\baseresults};
\addplot+[boxplot={box extend=0.1, draw position=3}, ColorRH2, solid, fill=ColorRH2!20, mark=x] table [col sep=comma, y=rejection_rate_rh_2_1] {\baseresults};
\addplot+[boxplot={box extend=0.1, draw position=4}, ColorMCVRP, solid, fill=ColorMCVRP!20, mark=x] table [col sep=comma, y=rejection_rate_mc_vrp] {\baseresults};

\end{axis}
\end{tikzpicture}
\caption{Distribution of rejection rates for the trained policy~$\pi^*$ with neural-network architectures based on MLP (\textcolor{ColorLegendOurRLMLP}{$\blacksquare$}), KAN (\textcolor{ColorLegendOurRL}{$\blacksquare$}), and CNN (\textcolor{ColorLegendOurRLConv}{$\blacksquare$}),
across 5 different episodes from the \textbf{microtransit data}.}
\Description{Distribution of rejection rates for the trained policy~$\pi^*$ with neural-network architectures based on MLP (\textcolor{ColorLegendOurRLMLP}{$\blacksquare$}), KAN (\textcolor{ColorLegendOurRL}{$\blacksquare$}), and CNN (\textcolor{ColorLegendOurRLConv}{$\blacksquare$}),
across 5 different episodes from the \textbf{microtransit data}.}
\label{fig:accept_or_reject_diff_arch_full}
\end{figure}

\pgfplotstableread[col sep=comma,]{data/nyc/base/base_results.csv}\baseresults

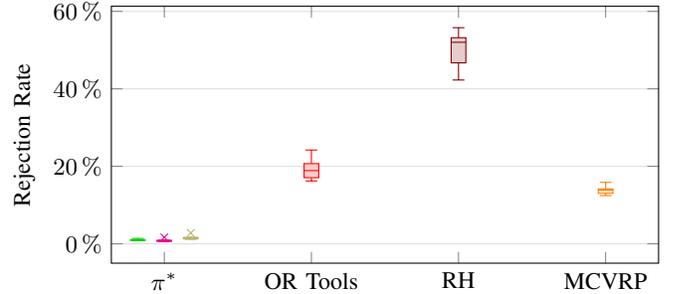
\begin{figure}
\begin{tikzpicture}
\begin{axis}[
      boxplot/draw direction=y,
      width=\columnwidth,
      xtick={1,2,3,4},
      xticklabel style={align=center},
      xticklabels={{$\pi^*$}, {OR Tools}, {RH}, {MCVRP}},
      height = 5.0cm,
      ymajorgrids,
      major grid style={draw=gray!25},
      bugsResolvedStyle/.style={},
      ylabel={Rejection Rate},
            yticklabel=\pgfmathprintnumber{\tick}\,$\%$,
      xlabel style={align=center},
      font=\small,
    ]

\addplot+[boxplot={box extend=0.1, draw position=1}, ColorOurRLMLP, solid,  lshift, fill=ColorOurRLMLP!20, mark=x] table [col sep=comma,y=rejection_rate_with_rejection_mlp] {\baseresults};
\addplot+[boxplot={box extend=0.1, draw position=1}, ColorOurRL, solid, fill=ColorOurRL!20, mark=x] table [col sep=comma, y=rejection_rate_with_rejection_kan] {\baseresults};
\addplot+[boxplot={box extend=0.1, draw position=1}, ColorOurRLConv, solid, rshift, fill=ColorOurRLConv!20, mark=x] table [col sep=comma,  y=rejection_rate_with_rejection_cnn] {\baseresults};
\addplot+[boxplot={box extend=0.1, draw position=2}, ColorORTools, solid, fill=ColorORTools!20, mark=x] table [col sep=comma, y=rejection_rate_rr_it] {\baseresults};
\addplot+[boxplot={box extend=0.1, draw position=3}, ColorRH2, solid, fill=ColorRH2!20, mark=x] table [col sep=comma, y=rejection_rate_rh_2_1] {\baseresults};
\addplot+[boxplot={box extend=0.1, draw position=4}, ColorMCVRP, solid, fill=ColorMCVRP!20, mark=x] table [col sep=comma, y=rejection_rate_mc_vrp] {\baseresults};

\end{axis}
\end{tikzpicture}
\caption{Distribution of rejection rates for the trained policy~$\pi^*$ with neural-network architectures based on MLP (\textcolor{ColorLegendOurRLMLP}{$\blacksquare$}), KAN (\textcolor{ColorLegendOurRL}{$\blacksquare$}), and CNN (\textcolor{ColorLegendOurRLConv}{$\blacksquare$}),
across 5 different episodes from the \textbf{NYC taxi data}.}
\Description{Distribution of rejection rates for the trained policy~$\pi^*$ with neural-network architectures based on MLP (\textcolor{ColorLegendOurRLMLP}{$\blacksquare$}), KAN (\textcolor{ColorLegendOurRL}{$\blacksquare$}), and CNN (\textcolor{ColorLegendOurRLConv}{$\blacksquare$}),
across 5 different episodes from the \textbf{NYC taxi data}.}
\label{fig:accept_or_reject_diff_arch_full_nyc}
\end{figure}

\cref{fig:accept_or_reject_diff_arch_full} shows the distribution of rejection rates for the trained policy $\pi^*$ with various neural-network architectures (MLP, KAN and CNN), evaluated over 5 different episodes from the microtransit data.
We observe that the MLP-based policy $\pi^*$ performs slightly better than policies based on KAN and CNN. 
Further, all three trained policies outperform all of the~baselines.

\cref{fig:accept_or_reject_diff_arch_full_nyc} shows the 
distribution of rejection rates for the trained policy $\pi^*$ with various neural-network architectures (MLP, KAN and CNN), evaluated over 5 different episodes from the NYC taxi data.
We again observe that the MLP-based trained policy $\pi^*$ performs slightly better than policies trained with KAN and CNN. Further, all three trained policies outperform all of the baselines. %

\subsection{Rejection Rate vs. Running Time for Anytime Algorithm with Various Neural-Network Architectures}
\label{app:varying_running_time}

\pgfplotstableread[col sep=comma,]{data/carta/search_time/search_time_cnn.csv}\convsearchtime

\pgfplotstableread[col sep=comma,]{data/carta/search_time/search_time_kan.csv}\kansearchtime

\pgfplotstableread[col sep=comma,]{data/carta/search_time/search_time_mlp.csv}\mlpsearchtime

\begin{figure}
\begin{tikzpicture}
\begin{axis}[
      boxplot/draw direction=y,
      width=\columnwidth,
      xtick={1,2,3,4,5,6,7},
      xticklabel style={align=center},
      xticklabels={{0}, {5}, {10}, {15 \\ Seconds}, {30}, {60}, {Full \\ Simulation}},
      height = 5.0cm,
      ymajorgrids,
      major grid style={draw=gray!25},
      bugsResolvedStyle/.style={},
      ylabel={Rejection Rate},
            yticklabel=\pgfmathprintnumber{\tick}\,$\%$,
      xlabel style={align=center},
      xlabel={},
      font=\small,
    ]

\addplot+[boxplot={box extend=0.1, draw position=1}, ColorOurRLConv, solid, lshift3,  fill=ColorOurRLConv!20, mark=x] table [col sep=comma, y=reject_insert_ar] {\convsearchtime};

\addplot+[boxplot={box extend=0.1, draw position=2}, ColorOurRLConv, solid, lshift3,  fill=ColorOurRLConv!20, mark=x] table [col sep=comma, y=reject_time_5_ar] {\convsearchtime};

\addplot+[boxplot={box extend=0.1, draw position=3}, ColorOurRLConv, solid,  lshift3, fill=ColorOurRLConv!20, mark=x] table [col sep=comma, y=reject_time_10_ar] {\convsearchtime};

\addplot+[boxplot={box extend=0.1, draw position=4}, ColorOurRLConv, solid,lshift3,  fill=ColorOurRLConv!20, mark=x] table [col sep=comma, y=reject_time_15_ar] {\convsearchtime};

\addplot+[boxplot={box extend=0.1, draw position=5}, ColorOurRLConv, solid,lshift3,  fill=ColorOurRLConv!20, mark=x] table [col sep=comma, y=reject_time_30_ar] {\convsearchtime};

\addplot+[boxplot={box extend=0.1, draw position=6}, ColorOurRLConv, solid, lshift3, fill=ColorOurRLConv!20, mark=x] table [col sep=comma, y=reject_time_60_ar] {\convsearchtime};

\addplot+[boxplot={box extend=0.1, draw position=7}, ColorOurRLConv, solid, lshift3,  fill=ColorOurRLConv!20, mark=x] table [col sep=comma, y=reject_search_ar] {\convsearchtime};

\addplot+[boxplot={box extend=0.1, draw position=1}, ColorOurRL, solid, fill=ColorOurRL!20, mark=x] table [col sep=comma, y=reject_insert_ar] {\kansearchtime};

\addplot+[boxplot={box extend=0.1, draw position=2}, ColorOurRL, solid, fill=ColorOurRL!20, mark=x] table [col sep=comma, y=reject_time_5_ar] {\kansearchtime};

\addplot+[boxplot={box extend=0.1, draw position=3}, ColorOurRL, solid, fill=ColorOurRL!20, mark=x] table [col sep=comma, y=reject_time_10_ar] {\kansearchtime};

\addplot+[boxplot={box extend=0.1, draw position=4}, ColorOurRL, solid, fill=ColorOurRL!20, mark=x] table [col sep=comma, y=reject_time_15_ar] {\kansearchtime};

\addplot+[boxplot={box extend=0.1, draw position=5}, ColorOurRL, solid, fill=ColorOurRL!20, mark=x] table [col sep=comma, y=reject_time_30_ar] {\kansearchtime};

\addplot+[boxplot={box extend=0.1, draw position=6}, ColorOurRL, solid, fill=ColorOurRL!20, mark=x] table [col sep=comma, y=reject_time_60_ar] {\kansearchtime};

\addplot+[boxplot={box extend=0.1, draw position=7}, ColorOurRL, solid, fill=ColorOurRL!20, mark=x] table [col sep=comma, y=reject_search_ar] {\kansearchtime};

\addplot+[boxplot={box extend=0.1, draw position=1}, ColorOurRLMLP, solid, rshift3, fill=ColorOurRLMLP!20, mark=x] table [col sep=comma, y=reject_insert_ar] {\mlpsearchtime};

\addplot+[boxplot={box extend=0.1, draw position=2}, ColorOurRLMLP, solid, rshift3, fill=ColorOurRLMLP!20, mark=x] table [col sep=comma, y=reject_time_5_ar] {\mlpsearchtime};

\addplot+[boxplot={box extend=0.1, draw position=3}, ColorOurRLMLP, solid, rshift3, fill=ColorOurRLMLP!20, mark=x] table [col sep=comma, y=reject_time_10_ar] {\mlpsearchtime};

\addplot+[boxplot={box extend=0.1, draw position=4}, ColorOurRLMLP, solid, rshift3, fill=ColorOurRLMLP!20, mark=x] table [col sep=comma, y=reject_time_15_ar] {\mlpsearchtime};

\addplot+[boxplot={box extend=0.1, draw position=5}, ColorOurRLMLP, solid, rshift3, fill=ColorOurRLMLP!20, mark=x] table [col sep=comma, y=reject_time_30_ar] {\mlpsearchtime};

\addplot+[boxplot={box extend=0.1, draw position=6}, ColorOurRLMLP, solid, rshift3, fill=ColorOurRLMLP!20, mark=x] table [col sep=comma, y=reject_time_60_ar] {\mlpsearchtime};

\addplot+[boxplot={box extend=0.1, draw position=7}, ColorOurRLMLP, solid, rshift3, fill=ColorOurRLMLP!20, mark=x] table [col sep=comma, y=reject_search_ar] {\mlpsearchtime};

\end{axis}
\end{tikzpicture}
\caption{Distribution of rejection rates for various running-time limits on the anytime algorithm  (in seconds, horizontal axis)
with trained policies $\pi^*$ based on 
MLP (\textcolor{ColorLegendOurRLMLP}{$\blacksquare$}), KAN (\textcolor{ColorLegendOurRL}{$\blacksquare$}), and CNN (\textcolor{ColorLegendOurRLConv}{$\blacksquare$}),
across 5 different episodes from the \textbf{microtransit data}.}
\Description{Distribution of rejection rates for various running-time limits on the anytime algorithm  (in seconds, horizontal axis)
with trained policies $\pi^*$ based on 
MLP (\textcolor{ColorLegendOurRLMLP}{$\blacksquare$}), KAN (\textcolor{ColorLegendOurRL}{$\blacksquare$}), and CNN (\textcolor{ColorLegendOurRLConv}{$\blacksquare$}),
across 5 different episodes from the \textbf{microtransit data}.}
\label{fig:all_search_time_accept_or_reject_decision}
\end{figure}
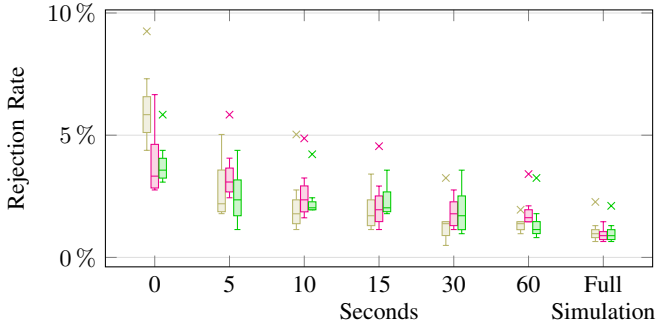

\pgfplotstableread[col sep=comma,]{data/nyc/search_time/search_time_cnn.csv}\convsearchtime

\pgfplotstableread[col sep=comma,]{data/nyc/search_time/search_time_kan.csv}\kansearchtime

\pgfplotstableread[col sep=comma,]{data/nyc/search_time/search_time_mlp.csv}\mlpsearchtime

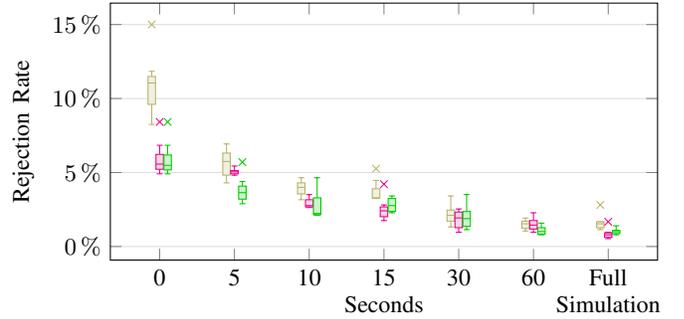
\begin{figure}[t]
\begin{tikzpicture}
\begin{axis}[
      boxplot/draw direction=y,
      width=\columnwidth,
      xtick={1,2,3,4,5,6,7},
      xticklabel style={align=center},
      xticklabels={{0}, {5}, {10}, {15 \\ Seconds}, {30}, {60}, {Full \\ Simulation}},
      height = 5.0cm,
      ymajorgrids,
      major grid style={draw=gray!25},
      bugsResolvedStyle/.style={},
      ylabel={Rejection Rate},
            yticklabel=\pgfmathprintnumber{\tick}\,$\%$,
      xlabel style={align=center},
      xlabel={},
      font=\small,
    ]

\addplot+[boxplot={box extend=0.1, draw position=1}, ColorOurRLConv, solid, lshift3,  fill=ColorOurRLConv!20, mark=x] table [col sep=comma, y=reject_insert_ar] {\convsearchtime};

\addplot+[boxplot={box extend=0.1, draw position=2}, ColorOurRLConv, solid, lshift3,  fill=ColorOurRLConv!20, mark=x] table [col sep=comma, y=reject_time_5_ar] {\convsearchtime};

\addplot+[boxplot={box extend=0.1, draw position=3}, ColorOurRLConv, solid,  lshift3, fill=ColorOurRLConv!20, mark=x] table [col sep=comma, y=reject_time_10_ar] {\convsearchtime};

\addplot+[boxplot={box extend=0.1, draw position=4}, ColorOurRLConv, solid,lshift3,  fill=ColorOurRLConv!20, mark=x] table [col sep=comma, y=reject_time_15_ar] {\convsearchtime};

\addplot+[boxplot={box extend=0.1, draw position=5}, ColorOurRLConv, solid,lshift3,  fill=ColorOurRLConv!20, mark=x] table [col sep=comma, y=reject_time_30_ar] {\convsearchtime};

\addplot+[boxplot={box extend=0.1, draw position=6}, ColorOurRLConv, solid, lshift3, fill=ColorOurRLConv!20, mark=x] table [col sep=comma, y=reject_time_60_ar] {\convsearchtime};

\addplot+[boxplot={box extend=0.1, draw position=7}, ColorOurRLConv, solid, lshift3,  fill=ColorOurRLConv!20, mark=x] table [col sep=comma, y=reject_search_ar] {\convsearchtime};

\addplot+[boxplot={box extend=0.1, draw position=1}, ColorOurRL, solid, fill=ColorOurRL!20, mark=x] table [col sep=comma, y=reject_insert_ar] {\kansearchtime};

\addplot+[boxplot={box extend=0.1, draw position=2}, ColorOurRL, solid, fill=ColorOurRL!20, mark=x] table [col sep=comma, y=reject_time_5_ar] {\kansearchtime};

\addplot+[boxplot={box extend=0.1, draw position=3}, ColorOurRL, solid, fill=ColorOurRL!20, mark=x] table [col sep=comma, y=reject_time_10_ar] {\kansearchtime};

\addplot+[boxplot={box extend=0.1, draw position=4}, ColorOurRL, solid, fill=ColorOurRL!20, mark=x] table [col sep=comma, y=reject_time_15_ar] {\kansearchtime};

\addplot+[boxplot={box extend=0.1, draw position=5}, ColorOurRL, solid, fill=ColorOurRL!20, mark=x] table [col sep=comma, y=reject_time_30_ar] {\kansearchtime};

\addplot+[boxplot={box extend=0.1, draw position=6}, ColorOurRL, solid, fill=ColorOurRL!20, mark=x] table [col sep=comma, y=reject_time_60_ar] {\kansearchtime};

\addplot+[boxplot={box extend=0.1, draw position=7}, ColorOurRL, solid, fill=ColorOurRL!20, mark=x] table [col sep=comma, y=reject_search_ar] {\kansearchtime};

\addplot+[boxplot={box extend=0.1, draw position=1}, ColorOurRLMLP, solid, rshift3, fill=ColorOurRLMLP!20, mark=x] table [col sep=comma, y=reject_insert_ar] {\mlpsearchtime};

\addplot+[boxplot={box extend=0.1, draw position=2}, ColorOurRLMLP, solid, rshift3, fill=ColorOurRLMLP!20, mark=x] table [col sep=comma, y=reject_time_5_ar] {\mlpsearchtime};

\addplot+[boxplot={box extend=0.1, draw position=3}, ColorOurRLMLP, solid, rshift3, fill=ColorOurRLMLP!20, mark=x] table [col sep=comma, y=reject_time_10_ar] {\mlpsearchtime};

\addplot+[boxplot={box extend=0.1, draw position=4}, ColorOurRLMLP, solid, rshift3, fill=ColorOurRLMLP!20, mark=x] table [col sep=comma, y=reject_time_15_ar] {\mlpsearchtime};

\addplot+[boxplot={box extend=0.1, draw position=5}, ColorOurRLMLP, solid, rshift3, fill=ColorOurRLMLP!20, mark=x] table [col sep=comma, y=reject_time_30_ar] {\mlpsearchtime};

\addplot+[boxplot={box extend=0.1, draw position=6}, ColorOurRLMLP, solid, rshift3, fill=ColorOurRLMLP!20, mark=x] table [col sep=comma, y=reject_time_60_ar] {\mlpsearchtime};

\addplot+[boxplot={box extend=0.1, draw position=7}, ColorOurRLMLP, solid, rshift3, fill=ColorOurRLMLP!20, mark=x] table [col sep=comma, y=reject_search_ar] {\mlpsearchtime};

\end{axis}
\end{tikzpicture}
\caption{Distribution of rejection rates for various running-time limits on the anytime algorithm  (in seconds, horizontal axis)
with trained policies $\pi^*$ based on 
MLP (\textcolor{ColorLegendOurRLMLP}{$\blacksquare$}), KAN (\textcolor{ColorLegendOurRL}{$\blacksquare$}), and CNN (\textcolor{ColorLegendOurRLConv}{$\blacksquare$}),
across 5 different episodes from the \textbf{NYC taxi data}.}
\Description{Distribution of rejection rates for various running-time limits on the anytime algorithm  (in seconds, horizontal axis)
with trained policies $\pi^*$ based on 
MLP (\textcolor{ColorLegendOurRLMLP}{$\blacksquare$}), KAN (\textcolor{ColorLegendOurRL}{$\blacksquare$}), and CNN (\textcolor{ColorLegendOurRLConv}{$\blacksquare$}),
across 5 different episodes from the \textbf{NYC taxi data}.}
\label{fig:all_search_time_accept_or_reject_decision_nyc}
\end{figure}

\cref{fig:all_search_time_accept_or_reject_decision,fig:all_search_time_accept_or_reject_decision_nyc} show how the rejection rate depends on the running time of the anytime algorithm, on microtransit and NYC taxi data, respectively.
The figures show results for trained policies $\pi^*$ based on various neural-network architectures (MLP, KAN, and CNN). 
The results of both figures demonstrate the importance of taking advantage of the time between request arrivals since the rejection rate decreases significantly as the running time increases, regardless of the dataset and the neural-network architecture.

\subsection{Fairness of Distributing Requests between Route Plans}
\label{app:evenness}

We assess the fairness of request distribution by analyzing the number of requests allocated to each driver (i.e., vehicle). Our findings indicate that the standard deviation of request distribution is approximately 14 when using Google OR-Tools, $\SimplePolicy$, and $\OptimalPolicy$, and around 13.5 with RH and MCVRP on the real-world microtransit dataset. Even though fairness is an important aspect, which ensures that all drivers are equally considered while serving requests, we primarily focus on maximizing the service rate in this work. As a result, requests may end up being distributed unevenly among~drivers.

\subsection{Ablation Study}
\label{app:ablation}

We compared our complete solution approach (i.e., use the learned $Q$-function for choosing the insertion placement, accept or reject decisions, as well as performing continual optimization) against different variations such as:

\begin{itemize}
\item [A.)] use the \textbf{simple heuristic objective} of maximizing idle time for the next 4 hours (i.e., policy $\SimplePolicy$) when making the insertion decision, as well as during the continual optimization via an anytime algorithm.
\item [B.)] use the \textbf{simple heuristic objective} (i.e., $\SimplePolicy$) to determine the insertion placement, while using the learned $Q$-function for making the insertion decision, as well as performing continual optimization.
\item [C.)] use the learned $Q$-function for choosing the insertion placement and the accept-or-reject decision, but use \textbf{simple heuristic objective} (i.e., $\SimplePolicy$) to perform continual optimization, always accepting requests if a feasible insertion exists.
\item [D.)] use the learned $Q$-function for choosing the insertion placement and the accept-or-reject decision, but use \textbf{simple heuristic objective} (i.e., $\SimplePolicy$)  to perform continual optimization.
\item [E.)] use the learned $Q$-function for making the insertion decision, as well as performing continual optimization, but always accept requests if a feasible insertion exists.
\end{itemize}

\begin{table*}[!ht]
    \centering
\begin{tabular}{ccccccc}
    \toprule
     \textbf{Settings} & \textbf{Insertion Decision} & \textbf{Acceptance Decision} & \textbf{Anytime Optimization} &  \textbf{Service Rate}  \\\midrule
     A & $\SimplePolicy$ & Always Accept ($\SimplePolicy$) & $\SimplePolicy$ & 98.64\\\hline
     B & $\SimplePolicy$ & Accept/Reject ($\OptimalPolicy$)& $\OptimalPolicy$   & 98.83 \\\hline
     C & $\OptimalPolicy$ & Always Accept ($\OptimalPolicy$) & $\SimplePolicy$   & 98.70  \\\hline
     D & $\OptimalPolicy$ & Accept/Reject ($\OptimalPolicy$)& $\SimplePolicy$ & 98.86 \\\hline
     E & $\OptimalPolicy$ & Always Accept ($\OptimalPolicy$) & $\OptimalPolicy$ & \textbf{99.06}\\\hline
     Our & $\OptimalPolicy$ & Accept/Reject ($\OptimalPolicy$)& $\OptimalPolicy$ & 98.83 \\\bottomrule
\end{tabular}
        \caption{Average service rate across 5 different episodes from \textbf{microtransit data} evaluated using the decision making choices while using learned $Q$-function using MLP based architecture.}
    \label{tab:ablation_mtd_mlp}
\end{table*}

\begin{table*}[!ht]
    \centering
\begin{tabular}{ccccccc}
    \toprule
     \textbf{Settings} & \textbf{Insertion Decision} & \textbf{Acceptance Decision} & \textbf{Anytime Optimization} &  \textbf{Service Rate}  \\\midrule
     A & $\SimplePolicy$ & Always Accept ($\SimplePolicy$) & $\SimplePolicy$ & 98.64\\\hline
     B & $\SimplePolicy$ & Accept/Reject ($\OptimalPolicy$)& $\OptimalPolicy$   & 98.25 \\\hline
     C & $\OptimalPolicy$ & Always Accept ($\OptimalPolicy$) & $\SimplePolicy$   &  98.67 \\\hline
     D & $\OptimalPolicy$ & Accept/Reject ($\OptimalPolicy$)& $\SimplePolicy$ & \textbf{98.80} \\\hline
     E & $\OptimalPolicy$ & Always Accept ($\OptimalPolicy$) & $\OptimalPolicy$ & 98.67\\\hline
     Our & $\OptimalPolicy$ & Accept/Reject ($\OptimalPolicy$)& $\OptimalPolicy$ & 98.77 \\\bottomrule
\end{tabular}
        \caption{Average service rate across 5 different episodes from \textbf{microtransit data} evaluated using the decision making choices while using learned $Q$-function using CNN based architecture.}
    \label{tab:ablation_mtd_cnn}
\end{table*}

\begin{table*}[!ht]
    \centering
\begin{tabular}{ccccccc}
    \toprule
     \textbf{Settings} & \textbf{Insertion Decision} & \textbf{Acceptance Decision} & \textbf{Anytime Optimization} &  \textbf{Service Rate}  \\\midrule
     A & $\SimplePolicy$ & Always Accept ($\SimplePolicy$) & $\SimplePolicy$ & 98.64\\\hline
     B & $\SimplePolicy$ & Accept/Reject ($\OptimalPolicy$)& $\OptimalPolicy$   &  98.44 \\\hline
     C & $\OptimalPolicy$ & Always Accept ($\OptimalPolicy$) & $\SimplePolicy$   &  98.57  \\\hline
     D & $\OptimalPolicy$ & Accept/Reject ($\OptimalPolicy$)& $\SimplePolicy$ & 98.80 \\\hline
     E & $\OptimalPolicy$ & Always Accept ($\OptimalPolicy$) & $\OptimalPolicy$ & 98.83\\\hline
     Our & $\OptimalPolicy$ & Accept/Reject ($\OptimalPolicy$)& $\OptimalPolicy$ & \textbf{98.99} \\\bottomrule
\end{tabular}
        \caption{Average service rate across 5 different episodes from \textbf{microtransit data} evaluated using the decision making choices while using learned $Q$-function using KAN based architecture.}
    \label{tab:ablation_mtd_kan}
\end{table*}

\cref{tab:ablation_mtd_mlp,tab:ablation_mtd_cnn,tab:ablation_mtd_kan}  show a comparison of the service rate using different ablation settings for the microtransit data, where the $Q$-function is learned using the neural network architectures of MLP, CNN, and KAN respectively. 
When using the $Q$-function learned with an MLP-based neural network architecture, the ablation setting ``E'' performs slightly better than our solution approach. Similarly, when using a CNN-based neural network architecture, the ablation setting ``D'' performs slightly better than our solution approach. On the other hand, when using a KAN-based neural network architecture, our proposed approach outperforms all other settings in the ablation study.

\begin{table*}[!ht]
    \centering
\begin{tabular}{ccccccc}
    \toprule
     \textbf{Settings} & \textbf{Insertion Decision} & \textbf{Acceptance Decision} & \textbf{Anytime Optimization} &  \textbf{Service Rate}  \\\midrule
     A & $\SimplePolicy$ & Always Accept ($\SimplePolicy$) & $\SimplePolicy$ & 98.70 \\\hline
     B & $\SimplePolicy$ & Accept/Reject ($\OptimalPolicy$)& $\OptimalPolicy$   &  98.95 \\\hline
     C & $\OptimalPolicy$ & Always Accept ($\OptimalPolicy$) & $\SimplePolicy$   &  99.00 \\\hline
     D & $\OptimalPolicy$ & Accept/Reject ($\OptimalPolicy$)& $\SimplePolicy$ &  \textbf{99.29} \\\hline
     E & $\OptimalPolicy$ & Always Accept ($\OptimalPolicy$) & $\OptimalPolicy$ & 98.84 \\\hline
     Our & $\OptimalPolicy$ & Accept/Reject ($\OptimalPolicy$)& $\OptimalPolicy$ & 98.93 \\\bottomrule
\end{tabular}
        \caption{Average service rate across 5 different episodes from \textbf{NYC taxi data} evaluated using the decision making choices while using learned $Q$-function using MLP based architecture.}
    \label{tab:ablation_nyc_mlp}
\end{table*}

\begin{table*}[!ht]
    \centering
\begin{tabular}{ccccccc}
    \toprule
     \textbf{Settings} & \textbf{Insertion Decision} & \textbf{Acceptance Decision} & \textbf{Anytime Optimization} &  \textbf{Service Rate}  \\\midrule
     A & $\SimplePolicy$ & Always Accept ($\SimplePolicy$) & $\SimplePolicy$ & 98.70 \\\hline
     B & $\SimplePolicy$ & Accept/Reject ($\OptimalPolicy$)& $\OptimalPolicy$   &  98.47 \\\hline
     C & $\OptimalPolicy$ & Always Accept ($\OptimalPolicy$) & $\SimplePolicy$   &  98.95 \\\hline
     D & $\OptimalPolicy$ & Accept/Reject ($\OptimalPolicy$)& $\SimplePolicy$ &   \textbf{99.02}\\\hline
     E & $\OptimalPolicy$ & Always Accept ($\OptimalPolicy$) & $\OptimalPolicy$ & 98.35 \\\hline
     Our & $\OptimalPolicy$ & Accept/Reject ($\OptimalPolicy$)& $\OptimalPolicy$ & 98.26 \\\bottomrule
\end{tabular}
        \caption{Average service rate across 5 different episodes from \textbf{NYC taxi data} evaluated using the decision making choices while using learned $Q$-function using CNN based architecture.}
    \label{tab:ablation_nyc_cnn}
\end{table*}

\begin{table*}[!ht]
    \centering
\begin{tabular}{ccccccc}
    \toprule
     \textbf{Settings} & \textbf{Insertion Decision} & \textbf{Acceptance Decision} & \textbf{Anytime Optimization} &  \textbf{Service Rate}  \\\midrule
     A & $\SimplePolicy$ & Always Accept ($\SimplePolicy$) & $\SimplePolicy$ & 98.70 \\\hline
     B & $\SimplePolicy$ & Accept/Reject ($\OptimalPolicy$)& $\OptimalPolicy$   & 98.72  \\\hline
     C & $\OptimalPolicy$ & Always Accept ($\OptimalPolicy$) & $\SimplePolicy$   &  99.00 \\\hline
     D & $\OptimalPolicy$ & Accept/Reject ($\OptimalPolicy$)& $\SimplePolicy$ &  99.00 \\\hline
     E & $\OptimalPolicy$ & Always Accept ($\OptimalPolicy$) & $\OptimalPolicy$ & 98.84 \\\hline
     Our & $\OptimalPolicy$ & Accept/Reject ($\OptimalPolicy$)& $\OptimalPolicy$ & \textbf{99.05}  \\\bottomrule
\end{tabular}
        \caption{Average service rate across 5 different episodes from \textbf{NYC taxi data} evaluated using the decision making choices while using learned $Q$-function using KAN based architecture.}
    \label{tab:ablation_nyc_kan}
\end{table*}

\cref{tab:ablation_nyc_mlp,tab:ablation_nyc_cnn,tab:ablation_nyc_kan} show a comparison of the service rate using different ablation settings for the NYC taxi data, where the $Q$-function is learned using MLP, CNN, and KAN neural network architectures, respectively. When using a $Q$-function learned with MLP or CNN based neural network architectures, the ablation setting ``D'' performs slightly better than our solution approach. On the other hand, when using the KAN-based neural network architecture, our proposed approach outperforms all other settings in the ablation study.

\subsection{Abstract Setting}
\label{app:abstract_setting}

In the abstract setting, we represent the city as a simple 2D square grid (2500 $\times$ 2500). For simplicity, we model the request arrivals as a stationary Poisson process with an average arrival rate of 20 requests per every 3600 units of time. We model the distribution of the time difference between request arrival and requested pick-up time as an exponential distribution with a mean of 7200 units in length. We consider a vehicle capacity of $c = 8$ and let the number of vehicles be $|\Vehicles| = 4$. We calculate travel times as the Euclidean distance between two location coordinates. For the evaluation, we consider each episode to last 12 hours. Accordingly, we examine 5 different episodes with a total of 240 requests. To ensure a fair comparison, we evaluate the proposed approach and ablation settings on the same 5 sets of requests.

\begin{table*}[!ht]
    \centering
\begin{tabular}{ccccccc}
    \toprule
     \textbf{Settings} & \textbf{Insertion Decision} & \textbf{Acceptance Decision} & \textbf{Anytime Optimization} &  \textbf{Service Rate}  \\\midrule
     A & $\SimplePolicy$ & Always Accept ($\SimplePolicy$) & $\SimplePolicy$ & 56.67\\\hline
     B & $\SimplePolicy$ & Accept/Reject ($\OptimalPolicy$)& $\OptimalPolicy$   & 57.91 \\\hline
     C & $\OptimalPolicy$ & Always Accept ($\OptimalPolicy$) & $\SimplePolicy$   &  57.17 \\\hline
     D & $\OptimalPolicy$ & Accept/Reject ($\OptimalPolicy$)& $\SimplePolicy$ & 56.83 \\\hline
     E & $\OptimalPolicy$ & Always Accept ($\OptimalPolicy$) & $\OptimalPolicy$ & 58.17\\\hline
     Our & $\OptimalPolicy$ & Accept/Reject ($\OptimalPolicy$)& $\OptimalPolicy$ & \textbf{58.33} \\\bottomrule
\end{tabular}
        \caption{Average service rate across 5 different episodes from \textbf{abstract} environment evaluated using the decision making choices while using learned Q function using MLP based architecture.}
    \label{tab:ablation_abstract_mlp}
\end{table*}

\begin{table*}[!ht]
    \centering
\begin{tabular}{ccccccc} %
    \toprule
     \textbf{Settings} & \textbf{Insertion Decision} & \textbf{Acceptance Decision} & \textbf{Anytime Optimization} &  \textbf{Service Rate}  \\\midrule
     A & $\SimplePolicy$ & Always Accept ($\SimplePolicy$) & $\SimplePolicy$ & 56.67 \\\hline
     B & $\SimplePolicy$ & Accept/Reject ($\OptimalPolicy$)& $\OptimalPolicy$   & 59.08 \\\hline
     C & $\OptimalPolicy$ & Always Accept ($\OptimalPolicy$) & $\SimplePolicy$   &  56.84 \\\hline
     D & $\OptimalPolicy$ & Accept/Reject ($\OptimalPolicy$)& $\SimplePolicy$ & 57.92 \\\hline
     E & $\OptimalPolicy$ & Always Accept ($\OptimalPolicy$) & $\OptimalPolicy$ & 58.00 \\\hline
     Our & $\OptimalPolicy$ & Accept/Reject ($\OptimalPolicy$)& $\OptimalPolicy$ & \textbf{59.17} \\\bottomrule
\end{tabular}
        \caption{Average service rate across 5 different episodes from \textbf{abstract} environment evaluated using the decision making choices while using learned Q function using CNN based architecture.}
    \label{tab:ablation_abstract_cnn}
\end{table*}

\cref{tab:ablation_abstract_mlp,tab:ablation_abstract_cnn} compares the service rate across different ablation settings for abstract problem setting, where the $Q$-function is learned using MLP- and CNN-based neural network architectures, respectively. Our proposed approach outperforms all other settings in the ablation study when using the learned $Q$-function at all the decision-making points. Specifically, the $Q$-function learned using CNN-based neural network achieves a 2.5\% improvement over a simple heuristic policy (i.e., $\SimplePolicy$) instead of learned $Q$-function (i.e., ablation setting ``A'').

\subsection{Rejection Rate vs. Running Time for MC~VRP}
\label{app:mcvrp_performance}

\pgfplotstableread[col sep=comma,]{data/carta/base/base_results.csv}\baseresults

\begin{figure}
\begin{tikzpicture}
\begin{axis}[
      boxplot/draw direction=y,
      width=\columnwidth,
      xtick={1,2,3,4,5},
      xticklabel style={align=center},
      xticklabels={{1}, {5}, {10}, {30}, {60}},
      height = 5.0cm,
      ymajorgrids,
      major grid style={draw=gray!25},
      bugsResolvedStyle/.style={},
      ylabel={Rejection Rate},
            yticklabel=\pgfmathprintnumber{\tick}\,$\%$,
      xlabel={Compute Time per Decision Epoch [seconds]},
      xlabel style={align=center},
      ymin=0,
      font=\small,
    ]

\addplot+[boxplot={box extend=0.1, draw position=1}, ColorMCVRP, solid, fill=ColorMCVRP!20, mark=x] table [col sep=comma, y=rejection_rate_mc_vrp_1] {\baseresults};

\addplot+[boxplot={box extend=0.1, draw position=2}, ColorMCVRP, solid, fill=ColorMCVRP!20, mark=x] table [col sep=comma, y=rejection_rate_mc_vrp_5] {\baseresults};

\addplot+[boxplot={box extend=0.1, draw position=3}, ColorMCVRP, solid, fill=ColorMCVRP!20, mark=x] table [col sep=comma, y=rejection_rate_mc_vrp_10] {\baseresults};

\addplot+[boxplot={box extend=0.1, draw position=4}, ColorMCVRP, solid, fill=ColorMCVRP!20, mark=x] table [col sep=comma, y=rejection_rate_mc_vrp] {\baseresults};

\addplot+[boxplot={box extend=0.1, draw position=5}, ColorMCVRP, solid, fill=ColorMCVRP!20, mark=x] table [col sep=comma, y=rejection_rate_mc_vrp_60] {\baseresults};

\end{axis}
\end{tikzpicture}
\caption{Distribution of rejection rates when using MC VRP as the dynamic VRP solver with different running times per decision epoch,
across 5 different episodes from the \textbf{microtransit data}.}
\Description{Distribution of rejection rates when using MC VRP as the dynamic VRP solver with different running times per decision epoch,
across 5 different episodes from the \textbf{microtransit data}.}
\label{fig:mc_vrp_performance}
\end{figure}
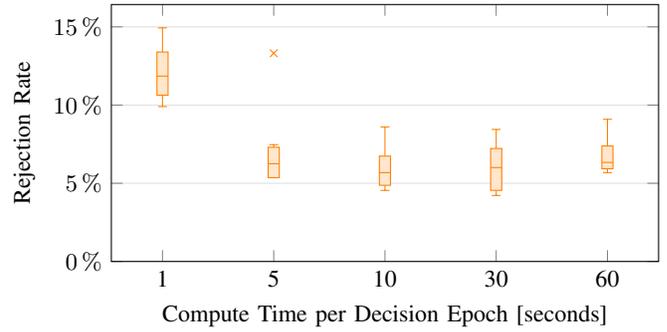

\pgfplotstableread[col sep=comma,]{data/nyc/base/base_results.csv}\baseresults

\begin{figure}
\begin{tikzpicture}
\begin{axis}[
      boxplot/draw direction=y,
      width=\columnwidth,
      xtick={1,2,3,4,5},
      xticklabel style={align=center},
      xticklabels={{1}, {5}, {10}, {30}, {60}},
      height = 5.0cm,
      ymajorgrids,
      major grid style={draw=gray!25},
      bugsResolvedStyle/.style={},
      ylabel={Rejection Rate},
            yticklabel=\pgfmathprintnumber{\tick}\,$\%$,
      xlabel={Compute Time per Decision Epoch [seconds]},
      xlabel style={align=center},
      ymin=0,
      font=\small,
    ]

\addplot+[boxplot={box extend=0.1, draw position=1}, ColorMCVRP, solid, fill=ColorMCVRP!20, mark=x] table [col sep=comma, y=rejection_rate_mc_vrp_1] {\baseresults};

\addplot+[boxplot={box extend=0.1, draw position=2}, ColorMCVRP, solid, fill=ColorMCVRP!20, mark=x] table [col sep=comma, y=rejection_rate_mc_vrp_5] {\baseresults};

\addplot+[boxplot={box extend=0.1, draw position=3}, ColorMCVRP, solid, fill=ColorMCVRP!20, mark=x] table [col sep=comma, y=rejection_rate_mc_vrp_10] {\baseresults};

\addplot+[boxplot={box extend=0.1, draw position=4}, ColorMCVRP, solid, fill=ColorMCVRP!20, mark=x] table [col sep=comma, y=rejection_rate_mc_vrp] {\baseresults};

\addplot+[boxplot={box extend=0.1, draw position=5}, ColorMCVRP, solid, fill=ColorMCVRP!20, mark=x] table [col sep=comma, y=rejection_rate_mc_vrp_60] {\baseresults};

\end{axis}
\end{tikzpicture}
\caption{Distribution of rejection rates when using MC VRP as the dynamic VRP solver with different running times per decision epoch,
across 5 different episodes from the \textbf{NYC taxi data}.}
\Description{Distribution of rejection rates when using MC VRP as the dynamic VRP solver with different running times per decision epoch,
across 5 different episodes from the \textbf{NYC taxi data}.}
\label{fig:mc_vrp_performance_nyc}
\end{figure}
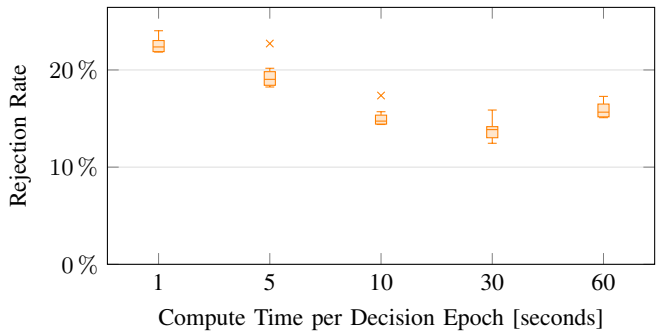

\cref{fig:mc_vrp_performance,fig:mc_vrp_performance_nyc} show how the rejection rate depends on the computation time at each decision epoch when using MC VRP as the dynamic VRP solver on the microtransit and NYC taxi datasets, respectively. 
We observe that as the time spent on each decision epoch increases, the rejection rate gradually decreases.

\subsection{Rejection Rate vs. Computation Time for Rolling Horizon}
\label{app:rh_performance}

\pgfplotstableread[col sep=comma,]{data/carta/base/base_results.csv}\baseresults

\begin{figure}
\begin{tikzpicture}
\begin{axis}[
      boxplot/draw direction=y,
      width=\columnwidth,
      xtick={1,2,3,4},
      xticklabel style={align=center},
      xticklabels={{1}, {5}, {10}, {15}},
      height = 5.0cm,
      ymajorgrids,
      major grid style={draw=gray!25},
      bugsResolvedStyle/.style={},
      ylabel={Rejection Rate},
            yticklabel=\pgfmathprintnumber{\tick}\,$\%$,
      xlabel={Maximum RTV generation time per batch \\in each decision epoch (in seconds)},
      xlabel style={align=center},
      font=\small,
    ]

\addplot+[boxplot={box extend=0.1, draw position=1}, ColorRH1, solid, fill=ColorRH1!20, mark=x] table [col sep=comma, y=rejection_rate_rh_1_1] {\baseresults};
\addplot+[boxplot={box extend=0.1, draw position=1}, ColorRH2, solid, rshift, fill=ColorRH2!20, mark=x] table [col sep=comma, y=rejection_rate_rh_2_1] {\baseresults};

\addplot+[boxplot={box extend=0.1, draw position=2}, ColorRH1, solid, fill=ColorRH1!20, mark=x] table [col sep=comma, y=rejection_rate_rh_1_5] {\baseresults};
\addplot+[boxplot={box extend=0.1, draw position=2}, ColorRH2, solid, rshift, fill=ColorRH2!20, mark=x] table [col sep=comma, y=rejection_rate_rh_2_5] {\baseresults};

\addplot+[boxplot={box extend=0.1, draw position=3}, ColorRH1, solid, fill=ColorRH1!20, mark=x] table [col sep=comma, y=rejection_rate_rh_1_10] {\baseresults};
\addplot+[boxplot={box extend=0.1, draw position=3}, ColorRH2, solid, rshift, fill=ColorRH2!20, mark=x] table [col sep=comma, y=rejection_rate_rh_2_10] {\baseresults};

\addplot+[boxplot={box extend=0.1, draw position=4}, ColorRH1, solid, fill=ColorRH1!20, mark=x] table [col sep=comma, y=rejection_rate_rh_1_15] {\baseresults};
\addplot+[boxplot={box extend=0.1, draw position=4}, ColorRH2, solid, rshift, fill=ColorRH2!20, mark=x] table [col sep=comma, y=rejection_rate_rh_2_15] {\baseresults};

\end{axis}
\end{tikzpicture}
\caption{Distribution of rejection rates when using Rolling Horizon  solver at each decision epoch
with rolling-horizon factors RH1 (\textcolor{ColorLegendRH1}{$\blacksquare$}) and RH2 (\textcolor{ColorLegendRH2}{$\blacksquare$}), across 5 different episodes from the \textbf{microtransit data}.}
\Description{Distribution of rejection rates when using Rolling Horizon  solver at each decision epoch
with rolling-horizon factors RH1 (\textcolor{ColorLegendRH1}{$\blacksquare$}) and RH2 (\textcolor{ColorLegendRH2}{$\blacksquare$}), across 5 different episodes from the \textbf{microtransit data}.}
\label{fig:rolling_horizon_performance}
\end{figure}
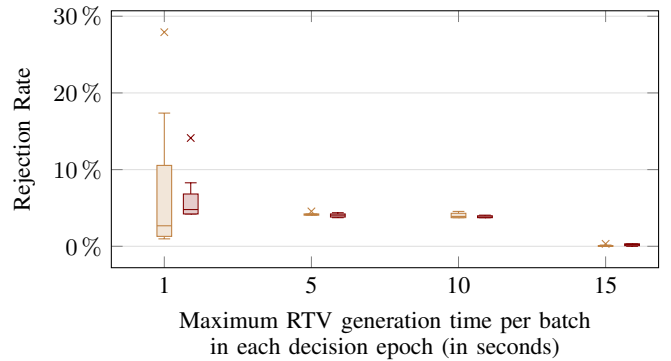

\pgfplotstableread[col sep=comma,]{data/carta/base/base_results.csv}\baseresults

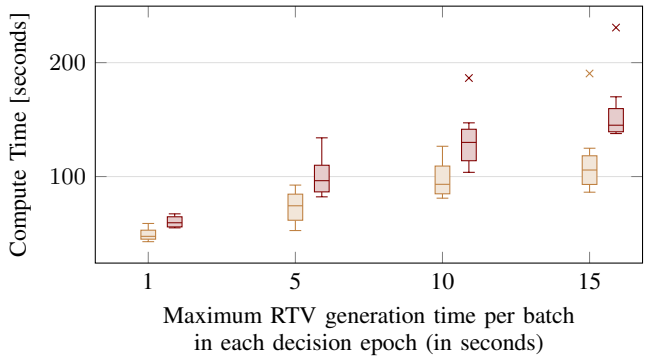
\begin{figure}
\begin{tikzpicture}
\begin{axis}[
      boxplot/draw direction=y,
      width=\columnwidth,
      xtick={1,2,3,4},
      xticklabel style={align=center},
      xticklabels={{1}, {5}, {10}, {15}},
      height = 5.0cm,
      ymajorgrids,
      major grid style={draw=gray!25},
      bugsResolvedStyle/.style={},
      ylabel={Compute Time [seconds]},
      xlabel={Maximum RTV generation time per batch \\in each decision epoch (in seconds)},
      xlabel style={align=center},
      font=\small,
    ]

\addplot+[boxplot={box extend=0.1, draw position=1}, ColorRH1, solid, fill=ColorRH1!20, mark=x] table [col sep=comma, y=compute_time_rh_1_1] {\baseresults};
\addplot+[boxplot={box extend=0.1, draw position=1}, ColorRH2, solid, rshift, fill=ColorRH2!20, mark=x] table [col sep=comma, y=compute_time_rh_2_1] {\baseresults};

\addplot+[boxplot={box extend=0.1, draw position=2}, ColorRH1, solid, fill=ColorRH1!20, mark=x] table [col sep=comma, y=compute_time_rh_1_5] {\baseresults};
\addplot+[boxplot={box extend=0.1, draw position=2}, ColorRH2, solid, rshift, fill=ColorRH2!20, mark=x] table [col sep=comma, y=compute_time_rh_2_5] {\baseresults};

\addplot+[boxplot={box extend=0.1, draw position=3}, ColorRH1, solid, fill=ColorRH1!20, mark=x] table [col sep=comma, y=compute_time_rh_1_10] {\baseresults};
\addplot+[boxplot={box extend=0.1, draw position=3}, ColorRH2, solid, rshift, fill=ColorRH2!20, mark=x] table [col sep=comma, y=compute_time_rh_2_10] {\baseresults};

\addplot+[boxplot={box extend=0.1, draw position=4}, ColorRH1, solid, fill=ColorRH1!20, mark=x] table [col sep=comma, y=compute_time_rh_1_15] {\baseresults};
\addplot+[boxplot={box extend=0.1, draw position=4}, ColorRH2, solid, rshift, fill=ColorRH2!20, mark=x] table [col sep=comma, y=compute_time_rh_2_15] {\baseresults};

\end{axis}
\end{tikzpicture}
\caption{Distribution of average computation time for each decision epoch
when using Rolling Horizon solver
with rolling-horizon factors RH1 (\textcolor{ColorLegendRH1}{$\blacksquare$}) and RH2 (\textcolor{ColorLegendRH2}{$\blacksquare$}), across 5 different episodes from the \textbf{microtransit data}.}
\Description{Distribution of average computation time for each decision epoch
when using Rolling Horizon solver
with rolling-horizon factors RH1 (\textcolor{ColorLegendRH1}{$\blacksquare$}) and RH2 (\textcolor{ColorLegendRH2}{$\blacksquare$}), across 5 different episodes from the \textbf{microtransit data}.}
\label{fig:rolling_horizon_time_performance}
\end{figure}

\cref{fig:rolling_horizon_performance,fig:rolling_horizon_time_performance} show how the Rolling Horizon solver's rejection rate and computation time (per decision epoch) vary with the running time for RTV generation per batch on the microtransit data. 
While allowing the RH solver to perform RTV generation for 15 seconds per batch could achieve very low rejection rates, 
the overall computation time per decision epoch (i.e., computation time to perform RTV generation for all batches and to solve the ILP problem) varies between 85 to 225 seconds. 
Running the RH solver for this long to confirm an incoming request is unacceptable in practical applications that aim to provide prompt confirmation.

\pgfplotstableread[col sep=comma,]{data/nyc/base/base_results.csv}\baseresults

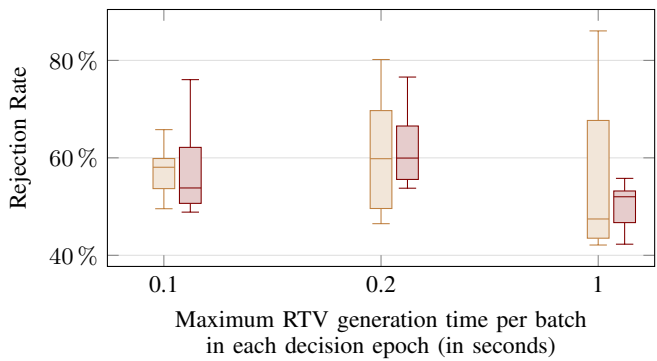
\begin{figure}
\begin{tikzpicture}
\begin{axis}[
      boxplot/draw direction=y,
      width=\columnwidth,
      xtick={1,2,3,4},
      xticklabel style={align=center},
      xticklabels={{0.1}, {0.2}, {1},{2}},
      height = 5.0cm,
      ymajorgrids,
      major grid style={draw=gray!25},
      bugsResolvedStyle/.style={},
      ylabel={Rejection Rate},
            yticklabel=\pgfmathprintnumber{\tick}\,$\%$,
      xlabel={Maximum RTV generation time per batch \\in each decision epoch (in seconds)},
      xlabel style={align=center},
      font=\small,
    ]

\addplot+[boxplot={box extend=0.1, draw position=1}, ColorRH1, solid, fill=ColorRH1!20, mark=x] table [col sep=comma, y=rejection_rate_rh_1_0_1] {\baseresults};
\addplot+[boxplot={box extend=0.1, draw position=1}, ColorRH2, solid, rshift, fill=ColorRH2!20, mark=x] table [col sep=comma, y=rejection_rate_rh_2_0_1] {\baseresults};

\addplot+[boxplot={box extend=0.1, draw position=2}, ColorRH1, solid, fill=ColorRH1!20, mark=x] table [col sep=comma, y=rejection_rate_rh_1_0_2] {\baseresults};
\addplot+[boxplot={box extend=0.1, draw position=2}, ColorRH2, solid, rshift, fill=ColorRH2!20, mark=x] table [col sep=comma, y=rejection_rate_rh_2_0_2] {\baseresults};

\addplot+[boxplot={box extend=0.1, draw position=3}, ColorRH1, solid, fill=ColorRH1!20, mark=x] table [col sep=comma, y=rejection_rate_rh_1_1] {\baseresults};
\addplot+[boxplot={box extend=0.1, draw position=3}, ColorRH2, solid, rshift, fill=ColorRH2!20, mark=x] table [col sep=comma, y=rejection_rate_rh_2_1] {\baseresults};

\end{axis}
\end{tikzpicture}
\caption{Distribution of rejection rates when using Rolling Horizon  solver at each decision epoch
with rolling-horizon factors RH1 (\textcolor{ColorLegendRH1}{$\blacksquare$}) and RH2 (\textcolor{ColorLegendRH2}{$\blacksquare$}), across 5 different episodes from the \textbf{NYC taxi data}.}
\Description{Distribution of rejection rates when using Rolling Horizon  solver at each decision epoch
with rolling-horizon factors RH1 (\textcolor{ColorLegendRH1}{$\blacksquare$}) and RH2 (\textcolor{ColorLegendRH2}{$\blacksquare$}), across 5 different episodes from the \textbf{NYC taxi data}.}
\label{fig:rolling_horizon_performance_nyc}
\end{figure}

\pgfplotstableread[col sep=comma,]{data/nyc/base/base_results.csv}\baseresults

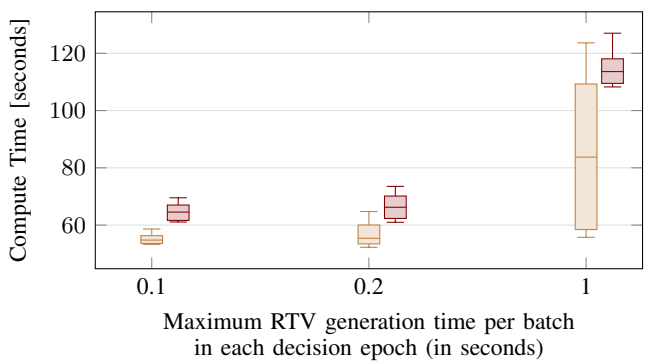
\begin{figure}
\begin{tikzpicture}
\begin{axis}[
      boxplot/draw direction=y,
      width=\columnwidth,
      xtick={1,2,3,4},
      xticklabel style={align=center},
      xticklabels={{0.1}, {0.2}, {1}, {2}},
      height = 5.0cm,
      ymajorgrids,
      major grid style={draw=gray!25},
      bugsResolvedStyle/.style={},
      ylabel={Compute Time [seconds]},
      xlabel={Maximum RTV generation time per batch \\in each decision epoch (in seconds)},
      xlabel style={align=center},
      font=\small,
    ]

\addplot+[boxplot={box extend=0.1, draw position=1}, ColorRH1, solid, fill=ColorRH1!20, mark=x] table [col sep=comma, y=compute_time_rh_1_0_1] {\baseresults};
\addplot+[boxplot={box extend=0.1, draw position=1}, ColorRH2, solid, rshift, fill=ColorRH2!20, mark=x] table [col sep=comma, y=compute_time_rh_2_0_1] {\baseresults};

\addplot+[boxplot={box extend=0.1, draw position=2}, ColorRH1, solid, fill=ColorRH1!20, mark=x] table [col sep=comma, y=compute_time_rh_1_0_2] {\baseresults};
\addplot+[boxplot={box extend=0.1, draw position=2}, ColorRH2, solid, rshift, fill=ColorRH2!20, mark=x] table [col sep=comma, y=compute_time_rh_2_0_2] {\baseresults};

\addplot+[boxplot={box extend=0.1, draw position=3}, ColorRH1, solid, fill=ColorRH1!20, mark=x] table [col sep=comma, y=compute_time_rh_1_1] {\baseresults};
\addplot+[boxplot={box extend=0.1, draw position=3}, ColorRH2, solid, rshift, fill=ColorRH2!20, mark=x] table [col sep=comma, y=compute_time_rh_2_1] {\baseresults};

\end{axis}
\end{tikzpicture}
\caption{Distribution of average computation time for each decision epoch
when using Rolling Horizon solver
with rolling-horizon factors RH1 (\textcolor{ColorLegendRH1}{$\blacksquare$}) and RH2 (\textcolor{ColorLegendRH2}{$\blacksquare$}), across 5 different episodes from the \textbf{NYC taxi data}.}
\Description{Distribution of average computation time for each decision epoch
when using Rolling Horizon solver
with rolling-horizon factors RH1 (\textcolor{ColorLegendRH1}{$\blacksquare$}) and RH2 (\textcolor{ColorLegendRH2}{$\blacksquare$}), across 5 different episodes from the \textbf{NYC taxi data}.}
\label{fig:rolling_horizon_time_performance_nyc}
\end{figure}

\cref{fig:rolling_horizon_performance_nyc,fig:rolling_horizon_time_performance_nyc} show how the Rolling Horizon solver's rejection rate and computation time (per decision epoch) vary with the running time for RTV generation per batch on the NYC taxi data. 
While spending more time on RTV generation per batch is able to reduce the rejection rate from 60\% to 40\%, the computation time increases drastically from 50 to 130 seconds per decision epoch.
Therefore, our proposed approach provides better performance both in terms of acceptance rate and in terms of time required to confirm a \hbox{request}.

\subsection{Statistical Tests of Significance}
Finally, we present the results of statistical tests on the significance of our experimental results shown by \cref{fig:accept_or_reject_full,fig:accept_or_reject_full_nyc}.
We performed paired two-sample permutation tests comparing the service rates of our proposed approach to the
service rates of each baseline approach on both datasets, based on 5 different episodes in each case. 
We observe that for both microtransit data and NYC data, the service rates obtained using our trained policy $\pi^*$ are significantly better than those of the baselines when we consider the significance threshold for the $p$-value to be 5\%.

\fi

\end{document}